\documentclass[10pt,twocolumn,letterpaper]{article}

\usepackage{iccv}
\usepackage{times}
\usepackage{epsfig}
\usepackage{graphicx}
\usepackage{amsmath}
\usepackage{amssymb}
\usepackage{bm}
\usepackage{verbatim}
\usepackage{booktabs}
\usepackage{enumitem}
\usepackage{authblk}
\usepackage[accsupp]{axessibility}
\usepackage[linesnumbered,ruled,vlined]{algorithm2e}
\usepackage{algorithmic}
\usepackage{appendix}

\usepackage[colorlinks=true,linkcolor=blue,citecolor=blue]{hyperref}
\iccvfinalcopy 

\def\iccvPaperID{2661} 

\ificcvfinal\fi

\begin{document}

\title{ASM: Adaptive Skinning Model for High-Quality 3D Face Modeling}
\author{Kai Yang, Hong Shang\thanks{Corresponding author.}, Tianyang Shi, Xinghan Chen, Jingkai Zhou, Zhongqian Sun, Wei Yang}
\affil{Tencent AI Lab}
\affil{{\tt\small \{arvinkyang, hongshang, tirionshi, xinghanchen, fszhou, sallensun, willyang\}@tencent.com}}

\maketitle
\ificcvfinal\thispagestyle{empty}\fi

\begin{abstract}
The research fields of parametric face model and 3D face reconstruction have been extensively studied. However, a critical question remains unanswered: how to tailor the face model for specific reconstruction settings. We argue that reconstruction with multi-view uncalibrated images demands a new model with stronger capacity. Our study shifts attention from data-dependent 3D Morphable Models (3DMM) to an understudied human-designed skinning model. We propose Adaptive Skinning Model (ASM), which redefines the skinning model with more compact and fully tunable parameters. With extensive experiments, we demonstrate that ASM achieves significantly improved capacity than 3DMM, with the additional advantage of model size and easy implementation for new topology. We achieve state-of-the-art performance with ASM for multi-view reconstruction on the Florence MICC Coop benchmark. Our quantitative analysis demonstrates the importance of a high-capacity model for fully exploiting abundant information from multi-view input in reconstruction. Furthermore, our model with physical-semantic parameters can be directly utilized for real-world applications, such as in-game avatar creation. As a result, our work opens up new research direction for parametric face model and facilitates future research on multi-view reconstruction.

\end{abstract}

\section{Introduction}

A key preliminary decision factor for 3D face modeling is a proper choice of face representation, as there is no one representation that fits all. For reconstruction with abundant constraints from multiple calibrated images (high-end), high capacity in the form of raw 3D points is essential to achieve high-fidelity scans with fine-grained details within the Multi-view Stereo (MVS) framework~\cite{beeler2010high, beeler2011high, furukawa2015multi, fyffe2017multi, li2021topologically}. For reconstruction with a single in-the-wild image (low-end), an intrinsically ill-posed problem, parametric face models with a strong prior are indispensable to ensure robust reconstruction with consistent topology~\cite{deng2019accurate, feng2021learning, zielonka2022towards, li2023robust}. Reconstruction with multi-view uncalibrated images (middle-end) is a previously less explored scenario with performance on par with the low-end setting, and far behind the high-fidelity scans in the high-end setting. This suggests that the additional constraints from multi-view uncalibrated images are not fully exploited. Previous studies in this category~\cite{amberg2007reconstructing, hernandez2017accurate, wu2019mvf, bai2020deep, bai2021riggable} have used parametric face models interchangeably with the low-end setting. We contend that parametric face models with a higher representation capacity should be employed to accommodate extra constraints from multi-view images. Consequently, this study investigates the design of high-capacity parametric face models for reconstruction with multi-view uncalibrated images. This understudied scenario is increasingly relevant in real-world applications due to the widespread use of high-quality camera-equipped mobile phones and the need for precise reconstruction for applications such as avatar creation and facial animation.

The parametric face model is an extensively researched field. The majority of studies are based on the 3D Morphable Model (3DMM), originally introduced in the pioneering work of Blanz and Vetter~\cite{blanz1999morphable}. Subsequent studies have continued to refine the 3DMM method by either improving the amount and diversity of data~\cite{li2017learning, yang2020facescape}  or proposing new methods~\cite{ranjan2018generating, tran2018nonlinear, bouritsas2019neural}  for dimensional reduction given such data. Simultaneously, a different trend has emerged in the game and film industries, where parametric face models are primarily represented in the form of human-designed skinning models. These models employ a set of controllable bones and skinning weights, which determine the degree to which each vertex on the mesh is influenced by the surrounding bones. This representation has demonstrated sufficient capability for extensive applications such as facial animation and avatar customization~\cite{james2005skinning, shi2019face, shi2020fast}. 

Comparing human-designed skinning models with data-dependent 3DMMs for 3D face modeling presents an intriguing yet understudied topic. These two models are fundamentally different in terms of constraint mechanism and capacity scaling. While 3DMMs derive their constraints from data, skinning models acquire proper constraints through the design process, such as converting empirical knowledge of real faces into the placement of bones and the definitions of skinning weights. Regarding capacity scaling, 3DMMs heavily rely on the collection of facial scan data, which is prohibitively expensive to scale. In contrast, the capacity of skinning models can be easily scaled by merely adjusting the number of parameters for bones and skinning weights, making it a more cost-effective and ideal candidate for high-capacity parametric face models.

With a closer look into standard skinning models with the vanilla Linear Blend Skinning (LBS), we find that their capacity can be further improved. Standard skinning models, which typically feature hundreds of bones on tens of thousands of vertices, usually posses tens of parameters for bone position, hundreds of parameters for transformation, and millions of parameters for skinning weights. These extensive skinning weights must be determined beforehand and remain fixed during subsequent 3D face modeling. They are usually determined either by professional animators or through data-driven learning~\cite{liu2019neuroskinning, pan2021heterskinnet}, with certain initial estimations~\cite{baran2007automatic}. Since skinning weights depend on bone position, which also needs to be predefined and fixed, transformation remains the sole variable in face modeling. Within this paradigm, improving model capacity relies on increasing the number of bones or refining predefined skinning weights. We refer to these standard skinning models as Static Skinning Models (SSM). We argue that the current paradigm of SSM fundamentally limits capacity, as the critical skinning weights are fixed. 

A neglected fact is that skinning weights, despite being defined in the form of a high-dimensional matrix, invariably result in low-dimensional patterns that are smooth, concentrated, and sparse. Given the strong structural nature of the human face, the movement space of each vertex is highly correlated and constrained. Consequently, skinning weights do not necessitate high-dimensional definition initially. We introduce the Adaptive Skinning Model (ASM), which defines skinning weights in a more compact form using the Gaussian Mixture Model (GMM). This new design significantly reduces the dimension of skinning weights to a level comparable with the transformation matrix. As a result, all parameters of skinning weights, transformation, and bone position can be simultaneously solved during reconstruction. This eliminates not only the labor-intensive manual design required in SSM but also the need for training data as in 3DMMs. Compared to SSM, our model can achieve a significantly increased capacity with even fewer total parameters. 

The main contributions of this paper are as follows:

\vspace{-2mm}
\begin{itemize}[itemsep=-1mm, leftmargin=*]
	\item A novel parametric face model is proposed, named ASM, by redefining skinning model with fully tunable parameters via introducing a more compact skinning weights representation with Gaussian Mixture Model.
	\item We demonstrate that ASM outperforms existing models in terms of capacity, model size, ease of implementation with arbitrary topology, and manual editing with semantic parameters. Moreover, it eliminates the need for laborious manual design and costly training data collection.
	\item State-of-the-art performance in 3D face reconstruction with multi-view uncalibrated images is achieved using ASM.

\end{itemize}

\section{Related Work}
{\noindent \bf 3D Morphable Models} was first proposed by Blanz and Vetter~\cite{blanz1999morphable} as a parametric face model. They used Principal Component Analysis (PCA) to reduce a set of topology-consistent face mesh into a low-dimensional space as a set of basis representing facial shape and texture. Paysan~\etal~\cite{paysan20093d} introduced Basel Face Model (BFM), which is a widely used 3DMM in recent years, calculated from registered 3D scans from 100 male and 100 female faces. FLAME~\cite{li2017learning} became popular recently, which used 3,800 face scans to construct a shape basis and 33,000 scans to construct the expression basis. FaceScape~\cite{yang2020facescape} collected high-quality facial data of 938 individuals and each with 20 expressions to build 3DMM with the bilinear PCA method.

To further improve the representation capacity of 3DMM, increasing attention has been drawn into non-linear dimensionality reduction methods, especially using neural networks to train and reduce facial library to latent vector features ~\cite{ranjan2018generating, tran2018nonlinear, bouritsas2019neural, zheng2022imface}. Ranjan~\etal~\cite{ranjan2018generating} introduced CoMA to extract the latent vector features from the mesh using an encoder-decoder network structure, resulting in better representations of the mesh from the training sets. Zheng~\etal~\cite{zheng2022imface} proposed ImFace, which used Signed Distance Function (SDF) and implicit neural representation to model human faces, achieving impressive results. Nevertheless, either linear or non-linear 3DMM methods are data dependent, making these methods intrinsically difficult to generalize and scale, considering collecting a large number of high-quality 3D facial models is prohibitively expensive.

{\noindent \bf Skinning Model} has a group of bones placed in 3D space, which can be controlled by the bones' translation, rotation, and scaling parameters. Once binding the bones with a mesh by defining the vertex-bone skinning weights matrix, the mesh can be deformed together with the bones via LBS. Skinning models have human-friendly semantic parameters, enabling the easy human design of bone placement and skinning weights. Besides, these models do not need to store basis and are computationally efficient. With these advantages, skinning models are widely used in the game and film industry for character modeling and animation of whole body and face.

Although popular in the game industry, skinning models receive less attention in 3D face modeling research. JNR~\cite{vesdapunt2020jnr} is the closest study to ours, which modeled face shape entirely by a skinning model with 52 bones and learned skinning weights. To the best of our knowledge, JNR is the only previous study that applied skinning models for face registration and reconstruction. Our study differs substantially from JNR in terms of design concepts and experimental findings. Firstly, JNR reduced the skinning weight matrix using a neural network, while we redesign the skinning model in a compact form in the first place, so that further dimension reduction or data-dependent learning are completely avoided, and all the parameters of skinning weights and bone positions can be freely solved online. Secondly, JNR demonstrated that skinning models achieved slightly worse capacity than state-of-the-art (SOTA) methods, such as FLAME, while our model achieves SOTA performance for both capacity and multi-view reconstruction.

\begin{figure*}[tbp]
	\begin{center}
		\includegraphics[width=0.95\textwidth]{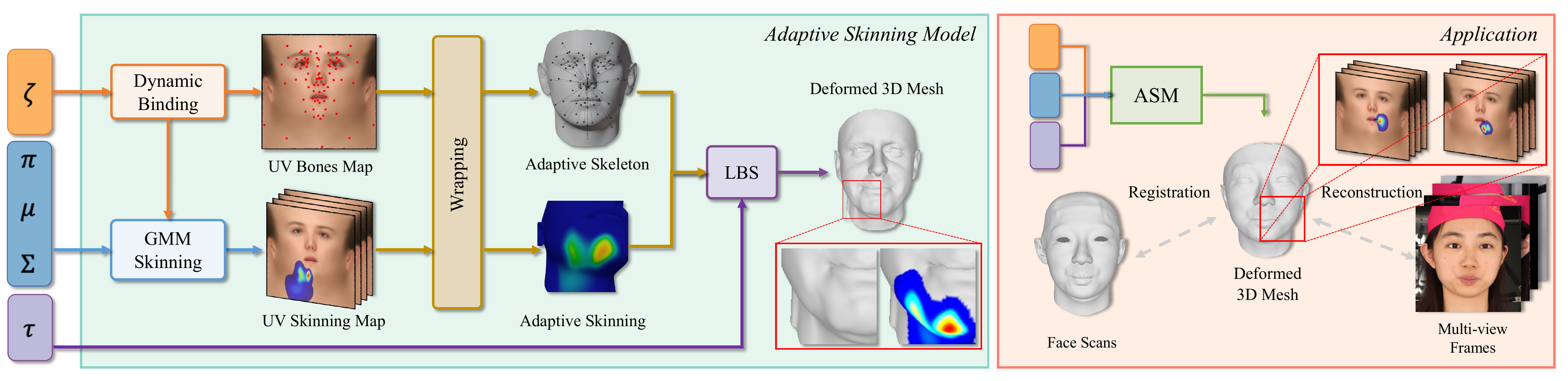}
	\end{center}
	\caption{Illustration of Adaptive Skinning Model. The bone positions in the UV space are adjusted by the parameters $\bm{\zeta}$, which also provide an initial guess for the GMM skinning module. The parameters $\bm{\pi}, \bm{\mu}, \bm{\Sigma}$ generate personal-specific skinning weights for each bone in the UV space, which is then wrapped into 3D space to obtain the updated skinning model. The output 3D mesh is deformed using LBS with the parameters $\bm{\tau}$. ASM can be used for tasks such as multi-view reconstruction and scan registration.}
	\label{fig2}
\end{figure*}

\section{Method}
In this section, we will begin by providing a brief overview of LBS, followed by an introduction of our proposed Adaptive Skinning Model (ASM). 

\subsection{Linear Blend Skinning}
LBS is a fundamental algorithm used for skeletal shape deformation in computer graphics~\cite{kavan2014skinning}. It requires three types of input data: vertex data from a polygon mesh, bone transformation data in the skeleton, and skinning weight data that defines the influence of each bone on each vertex. Given a vertex $\mathbf{v} \in \mathbb{R}^3$, the LBS algorithm computes its deformed position $\mathbf{v}'$ as follows:
\begin{equation}
	\mathbf{v}' = \sum^J_{j=1} w_j \mathbf{T}_j \mathbf{v}
	\label{lbs}
\end{equation}
where $\mathbf{v}$ and $\mathbf{v}'$ are in homogeneous coordinate format, $w_j$ is the skinning weight of bone $j$ on vertex $\mathbf{v}$ with the constraint $\sum^J_{j=1} w_j = 1$, $\mathbf{T}_j \in \mathbb{R}^{4\times4}$ is the bone $j$'s transformation matrix and $J$ is the total number of bones. In Eq.~\ref{lbs}, the deformation is performed by $\mathbf{T}_j$ according to the following formula:
\begin{equation}
	\mathbf{T}_j = \mathbf{M}^{l2w}_j \mathbf{M}^{w2l}_j = \mathbf{M}^{l2w}_p \mathbf{M}^{trs}(\bm{\tau}_j)\mathbf{B}_j^{-1}
	\label{transformation}
\end{equation}
where the vertex $\mathbf{v}$ is firstly projected from world space to local bone space by world-to-local transformation matrix $\mathbf{M}^{w2l}_j$ and then projected back into world space using $\mathbf{M}^{l2w}_j$. $\mathbf{M}^{l2w}_j$ can be decomposed into its parent bone's transformation matrix $\mathbf{M}^{l2w}_p$ multiply its local transformation $\mathbf{M}^{trs}(\bm{\tau}_j)$, where transformation parameters $\bm{\tau} \in \mathbb{R}^9$ include the translation, rotation, and scale parameters of the bone and $\mathbf{M}^{trs}(\cdot)$ is the composite matrix of these transformation parameters. $\mathbf{M}^{w2l}_j$ is defined as the inverse of pre-calculated bind-pose matrix $\mathbf{B}_j\in \mathbb{R}^{4\times4}$.

Based on Eq.~\ref{lbs} and Eq.~\ref{transformation}, for the vanilla LBS-based skinning model, only transformation parameters $\bm{\tau}$ can be adjusted for deformation, while the skinning weights and initial bone position are fixed, which significantly limits its capacity.

\subsection{Adaptive Skinning Model}
To further enlarge the capacity of the vanilla LBS-based skinning model, we redesign its skinning weights and binding strategy by introducing GMM skinning weights and dynamic binding. The proposed ASM can be written as:
\begin{equation}
	\begin{split}
            \small
		\begin{aligned}
			&{ASM}(\mathbf{v} | \bm{\zeta}, \bm{\pi}, \bm{\mu}, \bm{\Sigma}, \bm{\tau}) =\\ &\sum^J_{j=1} W^{g}(\mathbf{v} | \bm{\zeta}_j, \bm{\pi}, \bm{\mu}, \bm{\Sigma})  \mathbf{M}^{l2w}_p \mathbf{M}^{trs}(\bm{\tau}_j)B_j(F'(\bm{\zeta}))^{-1} \mathbf{v}
			\label{f7}
		\end{aligned}
	\end{split}
\end{equation}
where $W^g(\cdot)$ denotes GMM skinning weight function. $B(\cdot)$ is no longer the pre-calculated bind-pose matrix, but the standard bind-pose calculation method which takes positions and orientation in the world space of all the bones as inputs and outputs the bind-pose for each bone. $F'(\cdot)$, $\bm{\zeta}$, $\bm{\pi}$, $\bm{\mu}$, $\bm{\Sigma}$, and $\bm{\tau}$ will be described in detail below.  Fig.~\ref{fig2} presents an overview of our proposed model.

\vspace{1mm}
\noindent\textbf{GMM Skinning Weights.} Observing that skinning weights painted by human artists resemble a mixture of multiple Gaussian distributions, we introduce GMM to simulate the hand-painting process, so that we can build a more compact representation while maintaining strong capacity. Specifically, we define skinning weights as 2D-GMM in the unwrapped UV space.

Given the vertex $\mathbf{v}_i$ on the polygon mesh, there is a known unwrapping function $\mathbf{u}_i = F(\mathbf{v}_i)$ that maps the topology of the mesh vertex index to the UV space coordinate $\mathbf{u}_i\in\mathbb{R}^2$. 
The skinning weight of the point on the UV space influenced by bone $j$ is:
\begin{equation}
	\begin{split}
		\begin{aligned}
			W(\mathbf{v} | \bm{\zeta}_j, \bm{\pi}, \bm{\mu}, \bm{\Sigma}) = \sum^K_{k=1} \pi_{k} \mathcal{N}(F(\mathbf{v})| \bm{\mu}_{k} + \bm{\zeta}_j, \bm{\Sigma}_{k}) 
		\end{aligned}
	\end{split}
	\label{gmm}
\end{equation}
where $\pi_{k}\in \mathbb{R}$ $(\sum^K_{k=1} \pi_{k} = 1)$, $\bm{\mu}_{k}\in \mathbb{R}^2$, $\bm{\Sigma}_{k}\in \mathbb{R}^3$ are the GMM parameters, and $K$ controls the complexity of GMM. Since $\bm{\Sigma}$ is a symmetric matrix, it has only 3 degrees of freedom. $\bm{\zeta}_j \in \mathbb{R}^2$ is the projection of the bone $j$ onto UV space, and we use this projection as an initial guess of GMM's center. To find this projection, we firstly project the bone $j$ with initial placement position $\bm{\psi}_j^0 \in \mathbb{R}^3 $ in 3D space along with the z-axis (i.e. front-view) and then search the nearest vertex with index $t$ as a proxy to obtain:
\begin{equation}
\bm{\zeta}_j = F(\mathbf{v}_t)
\label{nearest}
\end{equation}

For the LBS-based skinning model, all the skinning weights on vertex $\mathbf{v}$ have to add up to 1, thus we normalize 2D GMM-based skinning weights as below:
\begin{equation}
	W^g(\mathbf{v} | \bm{\zeta}_j, \bm{\pi}, \bm{\mu}, \bm{\Sigma}) = \frac{W(\mathbf{v} | \bm{\zeta}_j, \bm{\pi}, \bm{\mu}, \bm{\Sigma})}{\sum^J_{i=1} {W(\mathbf{v} | \bm{\zeta}_i, \bm{\pi}, \bm{\mu}, \bm{\Sigma})}}
	\label{wnormalize}
\end{equation}
where $J$ is the total number of bones. With this method, we can compress a large number of skinning weights into a few 2D GMM parameters.

\vspace{1mm}
\noindent\textbf{Dynamic Bone Binding.}
In the previous GMM skinning weights calculation, $\bm{\zeta}_j$ is the UV position of the predefined bone $j$. Taking these estimations as the initialization and jointly optimizing $\bm{\zeta}$ with skinning weights is a straightforward way to further increase model capacity. During the joint optimization process, the gradient not only comes from $W^g(\cdot)$, but also from the bind-pose calculation $B_j(F'(\bm{\zeta}))$, where $F'(\bm{\zeta}_j)$ should be a differentiable wrapping function that maps the given UV space coordinate $\bm{\zeta}_j$ to the corresponding 3D position $\bm{\psi}_j$. Here we define this wrapping function as follows:
\begin{equation}
	\begin{split}
		\begin{aligned}
	&\bm{\psi}_j = F'(\bm{\zeta}_j) = \alpha \mathbf{v}_A + \beta \mathbf{v}_B + \gamma \mathbf{v}_C - \mathbf{v}_{t} + \bm{\psi}_j^0 \\
        &\alpha, \beta, \gamma = Barycentric(\bm{\zeta}_j, \mathbf{u}_A, \mathbf{u}_B, \mathbf{u}_C)
        \end{aligned}
        \end{split}
	\label{delta_p}
\end{equation}
where $\alpha$, $\beta$ and $\gamma$ are the barycentric weights of $\bm{\zeta}_j$ with respect to the triangle $f_{ABC}$ which $\bm{\zeta}_j$ fall within. The vertices of triangle $f_{ABC}$ are $\mathbf{u}_A=F(\mathbf{v}_A)$, $\mathbf{u}_B=F(\mathbf{v}_B)$, and $\mathbf{u}_C=F(\mathbf{v}_C)$. $\mathbf{v}_t$ is the same vertex referred in Eq.~\ref{nearest} and $\bm{\psi}_j^0$ is the initial position of bone $j$. 

Once we wrap $\bm{\zeta}$ to the 3D position $\bm{\psi}$ by vertex interpolation, we can use $B(\bm{\psi})$ to calculate the updated bind-pose matrix and evaluate the loss subsequently. As the whole process is differentiable, $\bm{\zeta}$ can be joint optimized with GMM skinning weights using backpropagation. 

Up to this point, we achieve a fully parameterized representation of the LBS-based skinning model. The detailed proof process and formulas can be found in the supplemental materials.

\subsection{Implementation Details.}
To set up the initial placement of the bones, we use Blender\footnote{https://www.blender.org} and place $J=84$ bones with a hierarchical structure, which provides higher degrees of freedom than JNR~\cite{vesdapunt2020jnr}. We use Blender's automatic skinning weights generation method to obtain the initial skinning weights and fit our GMMs for initial parameters $\bm{\zeta}$, $\bm{\pi}$, $\bm{\mu}$, and $\bm{\Sigma}$. These parameters serve as the starting point for optimization when using ASM in reconstruction tasks. For different scenarios, we suggest using different $K$ values for the GMM model ($K = 2\sim5$). In total, each bone of ASM has $(11 + K*6)$ tunable parameters. The dimension counting is shown in Tab.~\ref{parameters}. 
\begin{table}[!htbp]
    \small
        \begin{center}
                \begin{tabular}{lccccc}
                        \toprule
                        Parameters  & $\bm{\zeta}$ & $\bm{\pi}$ & $\bm{\mu}$ & $\bm{\Sigma}$ & $\bm{\tau}$ \\
                        \midrule
                        Dimension  & $2$ & $K$ & $K*2$ & $K*3$ & $9$\\
                        \bottomrule
                \end{tabular}
        \end{center}
        \caption{Dimension of parameters for each bone.}
        \label{parameters}
\end{table}

\section{Experiments}
\subsection{Model Characteristics} \label{model capacity}
{\noindent \bf Representation capacity} of parametric face models was assessed by fitting the models to 3D face scans and measuring the scan-to-mesh error. We utilized the Adam optimizer in PyTorch~\cite{paszke2019pytorch} with a learning rate of 1e-3 and 300 iterations to solve the transformation parameters of rigid ICP and the model parameters as an optimization problem. Our error measurement adhered to the NoW-benchmark~\cite{sanyal2019learning} prototype and was confined to the same facial region for fair comparison among models with different face coverage. We used two publicly available datasets: the LYHM dataset~\cite{dai2020statistical}, which includes 1,212 scanned meshes of neutral faces with inconsistent topology, and a dataset from FaceScape~\cite{yang2020facescape}, with the same setting as ImFace~\cite{zheng2022imface}, containing 10 individuals with 20 different expressions per person, resulting in 200 total meshes with consistent topology. Note that FaceScape is not in a metrical space, hence the units of measurements on FaceScape are not in millimeters. 

\vspace{1mm}
\begin{table}[!htbp]
    \small
	\begin{center}
		\vspace{-0pt}
		\begin{tabular}{lcc}
			\toprule
			Methods & LYHM&FaceScape \\
			\midrule
			BFM~\cite{paysan20093d}& $0.372_{\pm0.163}$& $0.462_{\pm0.052}$ \\
			FLAME~\cite{li2017learning}& $0.246_{\pm0.072}$& $0.341_{\pm0.039}$ \\
			CoMA~\cite{ranjan2018generating} & $0.756_{\pm0.186}$& $1.088_{\pm0.162}$ \\
			FaceScape~\cite{yang2020facescape} & $0.341_{\pm0.185}$& $0.216_{\pm0.048}$ \\
			ImFace~\cite{zheng2022imface}& $0.339_{\pm0.119}$& $0.257_{\pm0.061}$ \\
			MetaHuman~\cite{games2021metahuman}& $0.234_{\pm0.089}$& $0.269_{\pm0.063}$ \\
			\midrule
			Ours& $\bm{0.228_{\pm0.072}}$& $\bm{0.210_{\pm0.025}}$ \\
			\bottomrule
			
		\end{tabular}
	\end{center}
	\caption{Scan-to-fitting error with the metric of 3D-Normalized Mean Error (NME) (mm for LYHM). (Lower is better)}
	\label{capacity-table}
\end{table}

\vspace{1mm}
\begin{figure}[htbp]
	\begin{center}
		\vspace{-0pt}
		\includegraphics[width=0.48\textwidth]{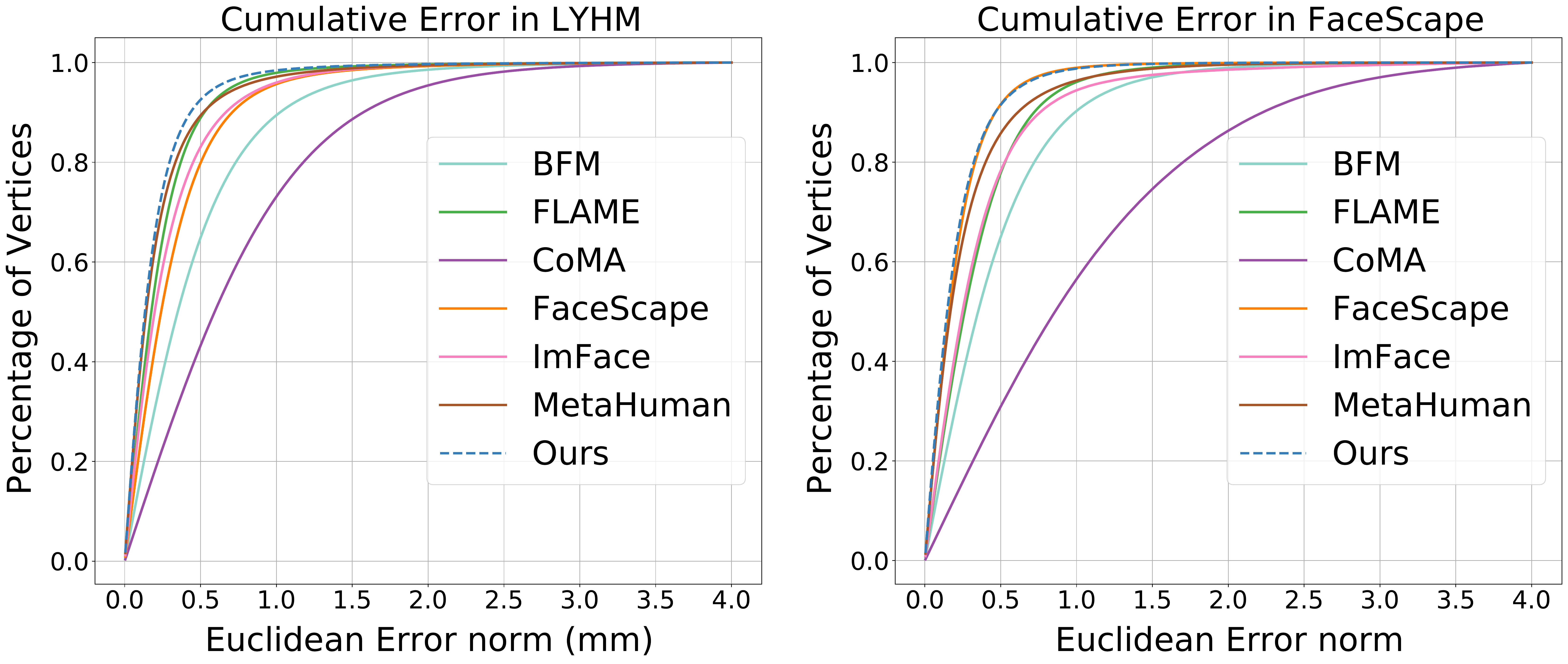}
	\end{center}
	\caption{Scan-to-fitting cumulative error curve. }
	\label{capacity-cumulative}
\end{figure}

\vspace{1mm}
\begin{figure}[tbp]
	\begin{center}
		\vspace{-0pt}
		\includegraphics[width=0.5\textwidth]{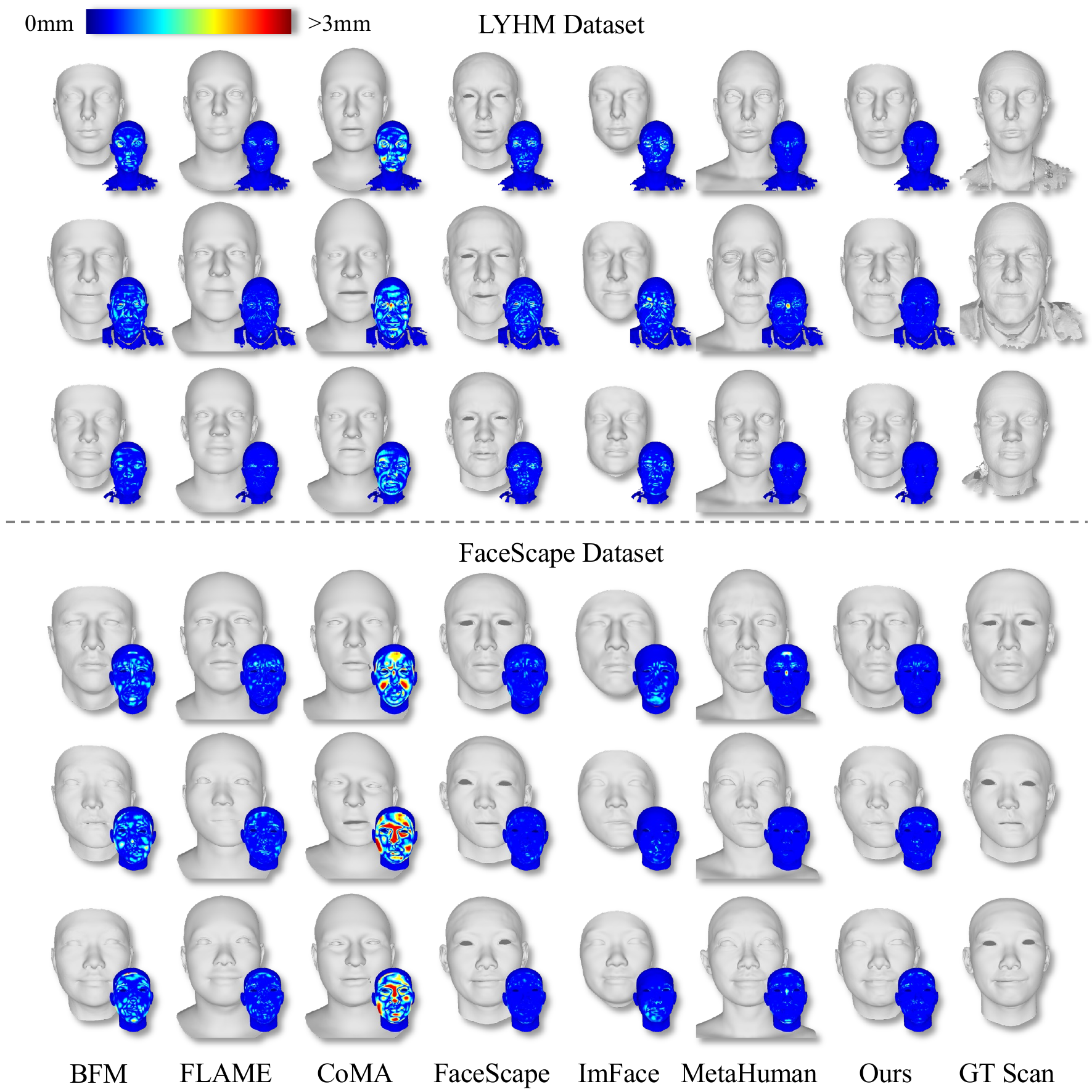}
	\end{center}
	\caption{Exemplar fitting result. GT Scans stands for the ground truth scan used for fitting.}
	\label{capacity-figure}
\end{figure}

The proposed ASM was compared to widely used and SOTA parametric face models, including BFM~\cite{paysan20093d} with entire 199 parameters of identity and 79 parameters of expression, FLAME~\cite{li2017learning} with entire 300 parameters of identity and 100 parameters of expression, FaceScape~\cite{yang2020facescape}, CoMA~\cite{ranjan2018generating}, and ImFace~\cite{zheng2022imface}. For CoMA, we used a 64-dimensional latent vector and retrained on its datasets, considering the original 8-dimensional latent vector would limit its performance. For nonlinear 3DMM (CoMA, ImFace) the latent vector served as parameters during fitting while the weights of the decoder network were fixed. Additionally, the state-of-the-art human-designed skinning model from MetaHuman Creator~\cite{games2021metahuman} was also compared, which included 887 bones, far more than our model. JNR~\cite{vesdapunt2020jnr} was not compared as its implementation and data were not open sourced.

Results in the form of mean error with standard deviation and cumulative error curve were shown in Tab.~\ref{capacity-table} and Fig.\ref{capacity-cumulative} respectively, with some examples shown in Fig.~\ref{capacity-figure}. Within the group of linear 3DMM, FLAME had the highest capacity and stable performance on both datasets. The extraordinary performance of FaceScape on its own dataset was illusive. When tested on a new dataset of LYHM, its performance dropped significantly, which illustrated the difficulty of generalization, a shared problem for all data-dependent methods. For non-linear 3DMM methods, CoMA had difficulty fitting these two datasets. ImFace behaved well on FaceScape datasets, but degraded on the LYHM dataset, similarly to FaceScape. Noted that both ImFace and FaceScape were trained using the FaceScape datasets, and both suffered from the generalization issue. Skinning models, including MetaHuman and our proposed ASM, though less studied previously, outperformed all data-dependent models. The intrinsic design of skinning models made it very cost-effective to increase capacity by simply adding more parameters. Compared to MetaHuman, the proposed ASM further improved capacity on both datasets with fewer tunable parameters, demonstrating the contribution of converting fixed skinning weights into compact and tunable skinning weights. Besides, skinning models avoided training data and the derived generalization issue, thus, leading to consistently excellent performance on both datasets. 

{\noindent \bf Implementation cost} is a practical consideration when adapting a face model to a new topology. It is common that different topologies are used by different groups in various applications. Off-the-shelf 3DMMs bring certain topologies, which may not be the desired ones in some applications. Adapting 3DMM to a new topology requires re-topologizing its data library and replicating the dimension reduction process, which is cumbersome for large-scale data as shown in Tab.~\ref{model_consumption_table}. It is even impossible if the data library is not accessible considering the risk of privacy. On the other hand, MetaHuman is a sophisticated human-designed SSM with 887 bones. Adapting MetaHuman to a new topology requires tremendous domain expertise and time-consuming painting of skinning weights.

\begin{figure}[htbp]
	\begin{center}
		\includegraphics[width=0.49\textwidth]{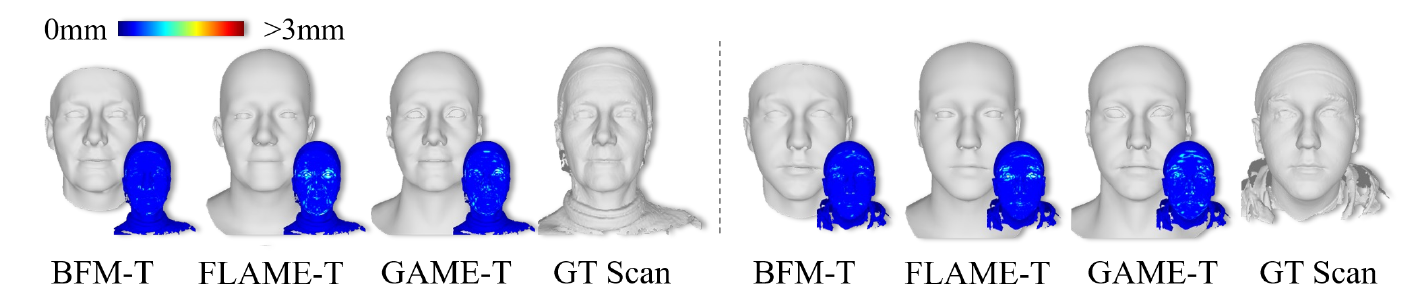}
	\end{center}
	\caption{Exemplar fitting results of ASM with different topologies. BFM-T, FLAME-T, and GAME-T stand for the topology of BFM, FLAME, and a game character respectively.}
	\label{model-topology-figure}
\end{figure}

In contrast, the implementation of our model is simply determining the number of bones and placing them on a facial mesh, which can be easily replicated on any new topology. For example, 84 bones were used in this work, which took around 20 minutes in total to go through the making process. As a demonstration, our original model with the topology from BFM was duplicated twice with the topology of FLAME and topology of a game character \footnote{We obtain the mesh file from the open game mods community: \\ https://steamcommunity.com/sharedfiles/filedetails/?id=2326367687}. Note the number and initial location of bones were kept the same among these three models. The representation capacity of these three models was tested on the LYHM datasets, with results shown in Tab.~\ref{model-topology-table} and some examples shown in Fig.~\ref{model-topology-figure}. Our method was robust for all different topologies. 

\vspace{1mm}
\begin{table}[!htbp]
    \small
	\begin{center}
		\vspace{-0pt}
        \addtolength{\tabcolsep}{-3pt}
		\begin{tabular}{lccc}
            \hline
            Topology & BFM & FLAME & GAME \\
            \hline
            3D-NME$\downarrow$ & $0.228_{\pm0.072}$ & $0.236_{\pm0.029}$ & $0.235_{\pm0.063}$ \\
            \hline
		\end{tabular}
	\end{center}
	\caption{Representation capacity of ASM with different topology. }
	\label{model-topology-table}
\end{table}

\vspace{1mm}

\begin{figure}[htbp]
	\begin{center}
		\includegraphics[width=0.45\textwidth]{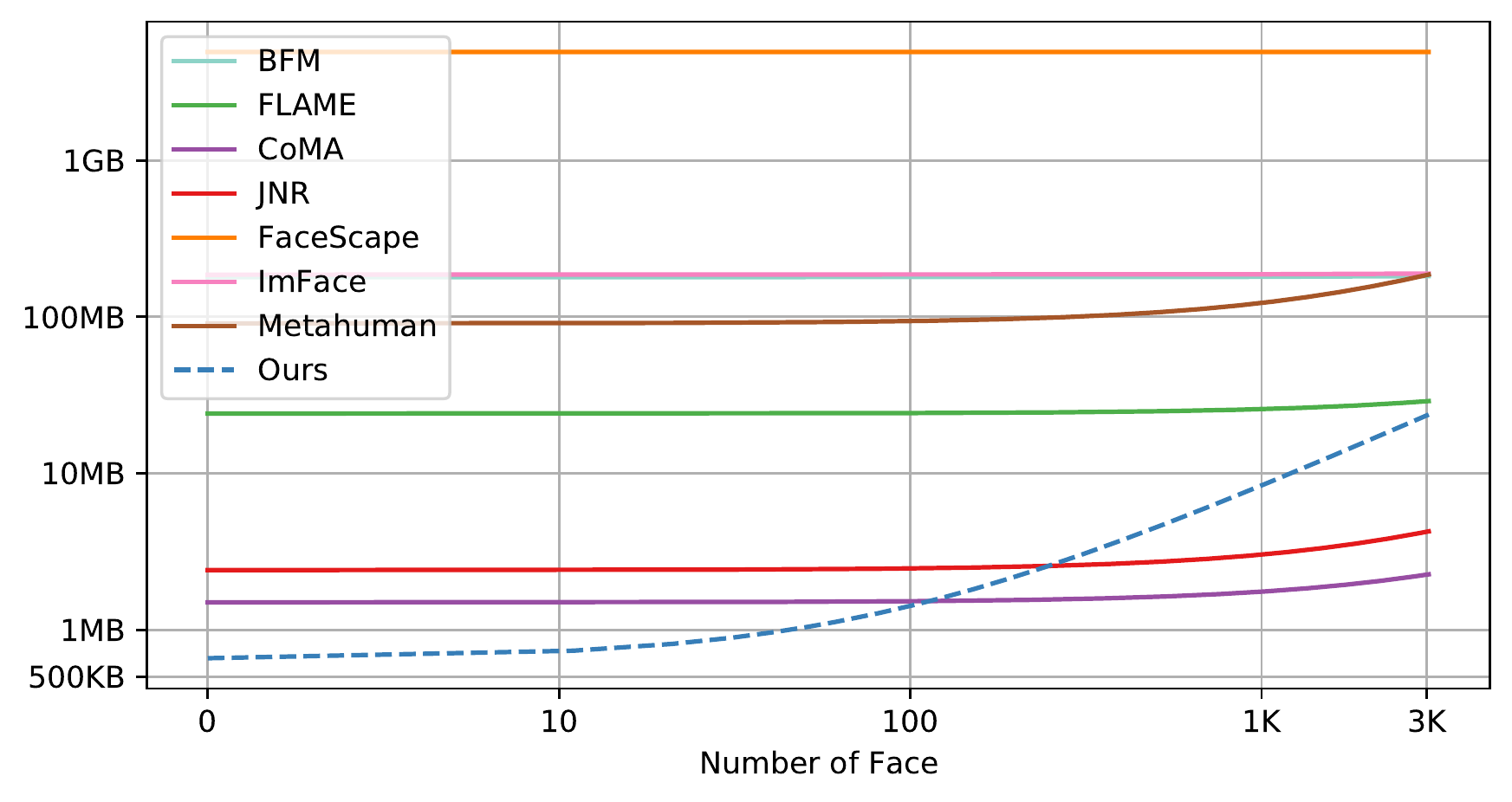}
	\end{center}
	\caption{Model size as a function for storing the number of faces.}
	\label{consumption-figure}
\end{figure}

{\noindent \bf Model size} refers to the disk space required to store the model, which is divided into the fixed part and headcount proportional part. The fixed part comes from the 3DMM basis, weights of neural networks, and predefined skinning weights. The headcount proportional part comes from 3DMM parameters, feature vectors of the neural networks, and skinning model tunable parameters. As shown in Fig.~\ref{consumption-figure}, our model size is significantly lower than all other models, especially within the range of 100 faces, which is a common range for real-world applications. This makes our model advantageous for mobile device applications.

\vspace{1mm}

\vspace{1mm}
\begin{table}[!htbp]
    \small
        \begin{center}
                \addtolength{\tabcolsep}{-3pt}
                \vspace{-0.cm}
                \begin{tabular}{lcccc}
                        \toprule
                        Methods& CPU& GPU & Dim. & Data \\
                        \midrule
                        BFM~\cite{paysan20093d}& 0.082s& 0.007s& 278 & 200 \\
                        FLAME~\cite{li2017learning}& 0.028s&0.002s& 406 & 3,800\\
                        CoMA~\cite{ranjan2018generating} & 1.880s&0.012s& 64 & 12\\
                        FaceScape~\cite{yang2020facescape} & 30.661s & 0.034s& 351 & 938\\
                        ImFace~\cite{zheng2022imface}& 94.660s & 20.816s& 256 & 355\\
                        MetaHuman~\cite{games2021metahuman}& 0.489s&0.007s& 7,983 & - \\
                        \midrule
                        Ours& 2.658s&0.066s& 1,932 & 1 \\
                        \bottomrule
                        
                \end{tabular}
        \end{center}
        \caption{Statistics of different face models. CPU and GPU refer to inference time measured on CPU or GPU. Dim refers to the dimension of parameters. Data refers to the number of individuals used to construct the face model.}
        \label{model_consumption_table}
\end{table}

{\noindent \bf Inference time} refers to the time it takes to generate a face mesh given the input parameters. Inference time measurement was conducted with a batch size of 32 and averaged over 1,000 repetitions. It was measured on either CPU of Intel(R) Xeon(R) Gold 6133 CPU @ 2.50GHz or the GPU of NVIDIA Tesla V100 32G. As shown in Tab.~\ref{model_consumption_table},  ASM was slower compared to linear 3DMM (BFM and FLAME) and SSM (Metahuman), but still within an acceptable range. ImFace with a much longer inference time increases the difficulty of being used.

\subsection{Model Application} \label{model application}
{\noindent \bf 3D face reconstruction} with multi-view uncalibrated images was evaluated. The Florence MICC benchmark is widely used for multi-view 3D face reconstruction with three subsets (Coop, indoor, and Outdoor). The Coop and Indoor subsets have video segments of 53 individuals with stable indoor lighting, differing by camera distance, portrait distance for Coop, and roof camera for Indoor. Coop is closer to our targeted setting with high-quality images, and both were used in our evaluation. For each video segment, we manually selected 15 frames at different angles with close expressions. Reconstruction was solved as an optimization problem with our proposed face model and photometric consistency constraints~\cite{amberg2007reconstructing, hernandez2017accurate}. A learning-based method~\cite{deng2019accurate} was used to serve as initialization to accelerate the convergence of optimization. For detailed experimental settings and energy function, please refer to the supplementary materials. As shown in the Tab.~\ref{MICC-recon}, We achieved SOTA performance on the Florence MICC Coop benchmark. For the Indoor benchmark with video taken in the distance, which is out of our targeted setting, methods with the advantage of robustness behave better, such as \cite{wood20223d}.

\vspace{1mm}
\begin{table}[!htbp]
    \small
	\begin{center}
		\begin{tabular}{lcc}
			\toprule
			Methods& Coop$\downarrow$& Indoor$\downarrow$ \\
			\midrule
			Piotraschke and Blanz~\cite{piotraschke2016automated}& $1.68$& $1.67$ \\
			Deng~\etal~\cite{deng2019accurate}& $1.60$& $1.61$\\
			Wood~\etal~\cite{wood20223d}& $1.43$& $\bm{1.42}$ \\
			Ours& $\bm{1.34}$& $1.53$ \\
			\bottomrule
			
		\end{tabular}
	\end{center}
	\caption{Multi-view reconstruction error with metric of 3D-RMSE(mm) on Florence MICC benchmark. ($\downarrow$Lower is better.)}
	\label{MICC-recon}
\end{table}

The MICC benchmark does not accurately represent our intended setting due to the allowance of speech and facial expression changes during video collection. To address this limitation, we conducted further evaluations on the FaceScape dataset, which captured a large number of high-definition images synchronously using a camera rig. Calibration information of this dataset was not used, and we randomly selected 3, 5, 10, and 20 images from 10 subjects to conduct multi-view 3D face reconstruction using various models, including BFM, FLAME, ASM-K2, ASM-K5, and MetaHuman, while maintaining consistent settings as previously stated. ASM-K2 and ASM-K5 referred to our model with different parameter $K$ settings, with ASM-K2 being the default setting used in all other experiments. Additionally, we also compared with MVS implemented by commercial software MetashapePro\footnote{https://www.agisoft.com}.

\begin{figure}[htbp]
	\begin{center}
		\vspace{-0.6cm}
		\includegraphics[width=0.49\textwidth]{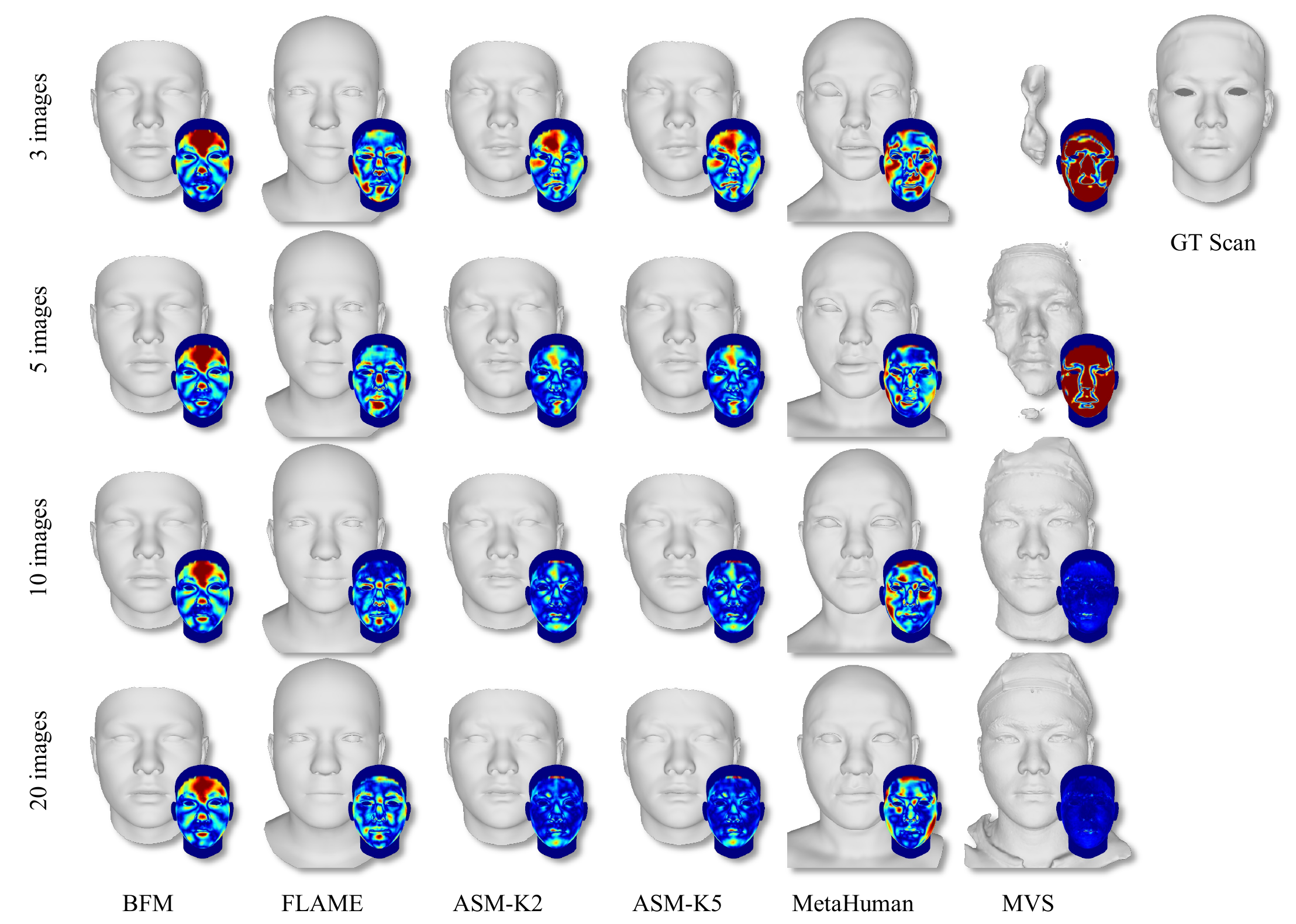}
	\end{center}
	\caption{Multi-view reconstruction result on FaceScape.}
	\label{facescape_recon_figure}
\end{figure}

Tab.~\ref{FaceScape-recon} and Fig.~\ref{facescape_recon_figure} demonstrated that skinning models, including ours and MetaHuman, outperform 3DMM (BFM and FLAME) in the multi-view setting. Skinning models can continuously improve results with more views, while 3DMM exhibited a less noticeable improvement. This highlighted the importance of using skinning models with higher capacity to accommodate more constraints from multi-view input. MVS failed with only 3 or 5 images, but achieved high-fidelity results with 20 images, as expected. While MetaHuman results exhibited bizarre shapes, our model achieved natural and high-fidelity results. This can be attributed to the fact that MetaHuman adds extra bones, far beyond the physical number of joints on human face. As a result, the added capacity may not align well with the actual human face, resulting in an unnatural appearance. In contrast, our proposed model increases capacity in a more balanced manner by allowing all skinning model parameters to be tuned simultaneously, leading to a better representation of human face.

\vspace{1mm}
\begin{table}[!htbp]
    \small
	\begin{center}
            \addtolength{\tabcolsep}{-4pt}
            \small
		\begin{tabular}{c|cccccc}
			\toprule
			Images& BFM& FLAME& ASM-K2 & ASM-K5 & MetaHuman & MVS \\
			\hline
			3 & $1.64$ & $1.56 $& $1.30 $& $\bm{1.29} $& $1.47$& - \\
			5 & $1.56$ & $1.54 $& $1.06 $& $\bm{1.06} $& $1.34$& - \\
			10 & $1.52$ & $1.48 $& $0.94 $& $0.92 $& $1.15 $&$ \bm{0.88}$ \\
			20 & $1.50$ & $1.33 $& $0.86 $& $0.84 $& $1.04 $& $\bm{0.55}$ \\
			\bottomrule
		\end{tabular}
	\end{center}
	\caption{Multi-view reconstruction error with metric of 3D-RMSE on selected FaceScape dataset. ($\downarrow$Lower is better.)}
	\label{FaceScape-recon}
\end{table}

\vspace{1mm}

\begin{figure}[htbp]
	\begin{center}
        \setlength{\belowcaptionskip}{-1cm}
        \includegraphics[width=0.48\textwidth]{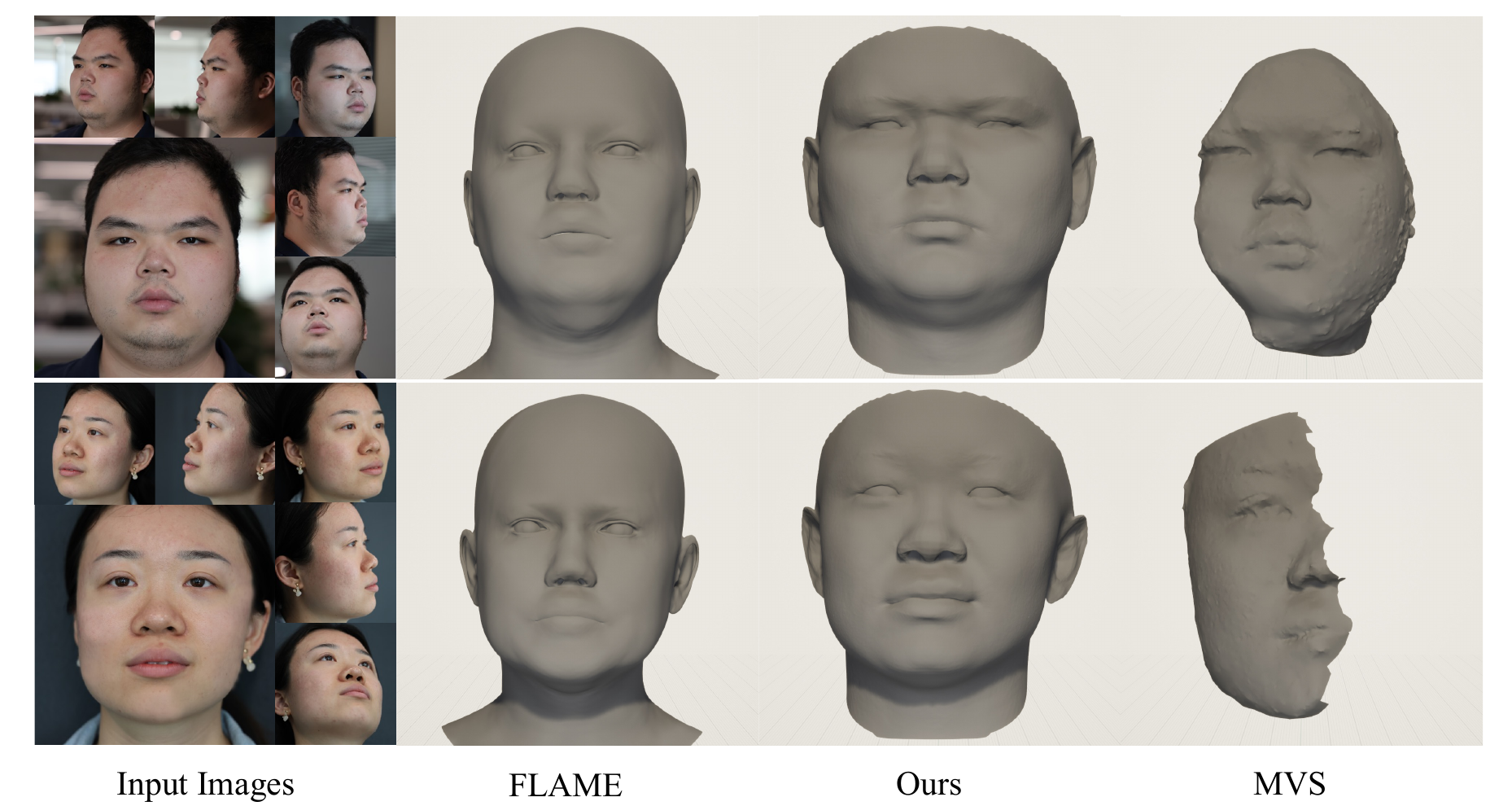}
	\end{center}
	\caption{Multi-view reconstruction result on in-house data.}
	\label{application_from_photos}
\end{figure}

We also obtained in-house data by capturing 6 images with a high-quality mobile camera and requesting participants to remain stationary. Using the same set-up, we performed multi-view 3D face reconstruction and compared our model to FLAME and MVS. Our model outperformed FLAME with more identifiable results, as shown in Fig.~\ref{application_from_photos}, while MVS failed. These findings demonstrate that our model is the proper choice for reconstruction with multi-view uncalibrated images, especially when the number of images is not adequate for successful MVS. 

\begin{figure}[htbp]
	\begin{center}
		\includegraphics[width=0.48\textwidth]{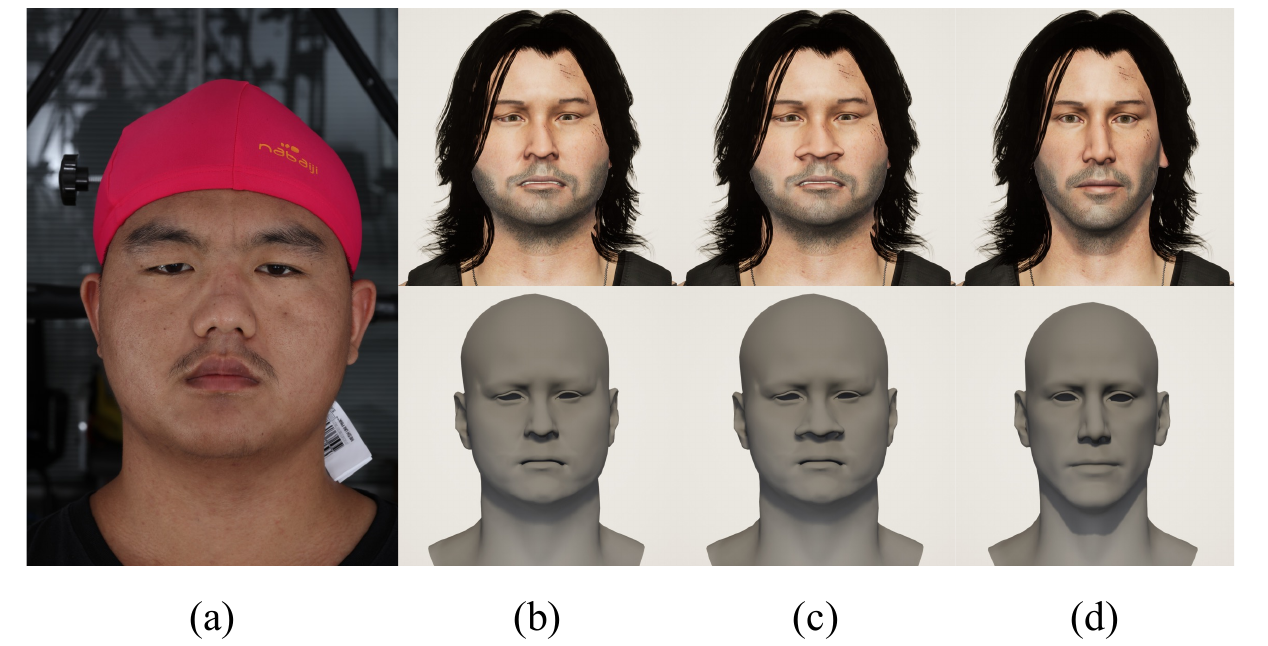}
	\end{center}
	\caption{(a) exemplar image out of 5; (b) customized avatar with reconstruction result; (c) avatar with further manual edit, for example, adjusting the bone of the nose wing; (d) the original avatar. }
	\label{fig3}
\end{figure}

{\noindent \bf In-game avatar creation} is another application benefiting from the proposed model, which is to customize in-game avatars given input images. Character's face is mostly represented in the form of skinning models with certain topology in games~\cite{shi2019face, shi2020fast}. Our model belongs to skinning models and can be easily adapted to new topology, therefore, the reconstruction results of our model can be directly transferred into the game system without a performance drop. The implementation of reconstruction had the same setting as above, except the model was based on the topology from the game, as previously illustrated in Fig.~\ref{model-topology-figure}.  As shown in Fig.~\ref{fig3}, in-game avatar from reconstruction result was achieved, and post-editing was allowed, due to the advantage of skinning models with physical-semantic parameters.

\subsection{Ablation Studies} 
A quantitative ablation study was conducted to investigate the key design components for fitting and reconstruction performance using the LYHM and MICC datasets, respectively. The study compared the following methods. SSM referred to a static skinning model with fixed bone binding and skinning weights provided by Blender. DBB referred to dynamic bone binding with the bone position as tunable variable. GSW referred to tunable GMM based skinning weights. RD referred to replacing the initial skinning weights provided by Blender with random ones. In Tab.~\ref{ablation_study_table}, the last row represented the default setting as in previous evaluations. Results indicated that SSM had a higher representation capacity than most 3DMM models, except FLAME, leading to improve multi-view reconstruction performance. Converting bone location and skinning weights into tunable parameters further improved capacity. Careful consideration was required for the initialization of GMM skinning weight.

\begin{table}[!htbp]
    \small
	\begin{center}
            \addtolength{\tabcolsep}{-1pt}
            \begin{tabular}{lccccc}
			\toprule 
			SSM& DBB & GSW & RD & Registration & Reconstruction \\
			\midrule
			$\checkmark$ & & & & $0.322_{\pm0.118}$ & $1.36_{\pm0.48}$ \\
			$\checkmark$ & $\checkmark$ & & & $0.282_{\pm0.094}$ & $1.36_{\pm0.46}$ \\
			$\checkmark$ & $\checkmark$ &$\checkmark$& $\checkmark$& $0.416_{\pm0.107}$ & $1.47_{\pm0.45}$ \\
			$\checkmark$ & $\checkmark$ & $\checkmark$& & $\bm{0.228_{\pm0.072}}$ & $\bm{1.34_{\pm0.51}}$ \\
			\bottomrule
			
		\end{tabular}
	\end{center}
	\caption{Ablation study on registration (with metric of 3D-NME) and reconstruction (with metric of 3D-RMSE).}
	\label{ablation_study_table}
\end{table}

\begin{figure}[htbp]
	\begin{center}
		\includegraphics[width=0.5\textwidth]{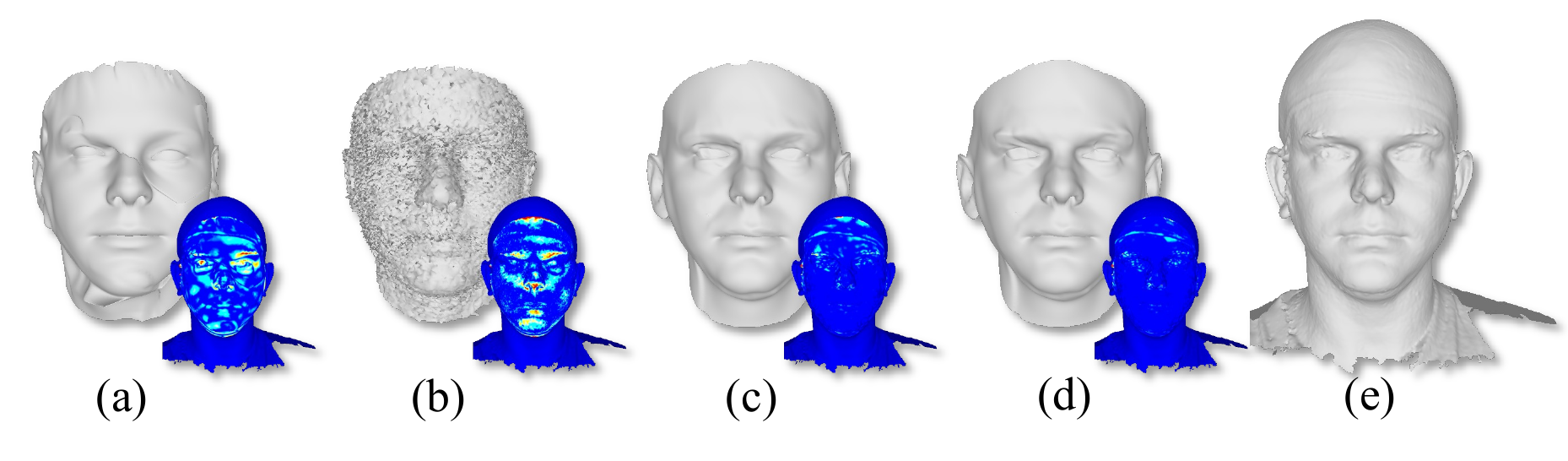}
	\end{center}
	\caption{Illustration of fitting results. (a) random bones placement; (b) random skinning weights; (c) enlarged regularization weight on  the eyebrow area; (d) default setting of ASM; (e) GT scans.}
	\label{regulazation}
\end{figure}

Additionally, a qualitative ablation study was conducted to illustrate the facial prior of ASM. Unlike 3DMMs learning constraints from data, ASM, as well as skinning models in general, encodes proper constraints within the design of initial bone placement and skinning weights. Therefore, random bone placement or skinning weights lead to failed modeling, as shown in Fig.~\ref{regulazation}(a) and Fig.~\ref{regulazation}(b) respectively. Besides, regularization terms can be used to provide additional constraints for local facial regions, thanks to the explicit semantics of skinning model parameters. For instance, enlarging the regularization weight for bones near the eyebrows can reduce the impact of noise for that region, as shown in Fig.~\ref{regulazation}(c).

\section{Discussion}
This study demonstrated that parametric face models have varying characteristics and should be tailored for specific applications. When dealing with low-quality input, such as the MICC Indoor benchmark, 3DMM with strong prior achieved robust and SOTA performance. For high-quality calibrated input captured within a camera rig, parametric face models were unnecessary, and MVS with raw 3D points achieved high-quality facial scans, considered as the ground truth. For intermediate-level input of high-quality multi-view but uncalibrated images, skinning models based ASM with higher capacity achieved SOTA performance on MICC Coop benchmark, FaceScape (without calibration), and our in-house data. Compared to a sophisticated human-designed static skinning model, ASM with fully tunable parameters can further improve capacity in a more natural and effective way.

\begin{figure}[tbp]
	\begin{center}
		\vspace{-0pt}
		\includegraphics[width=0.5\textwidth]{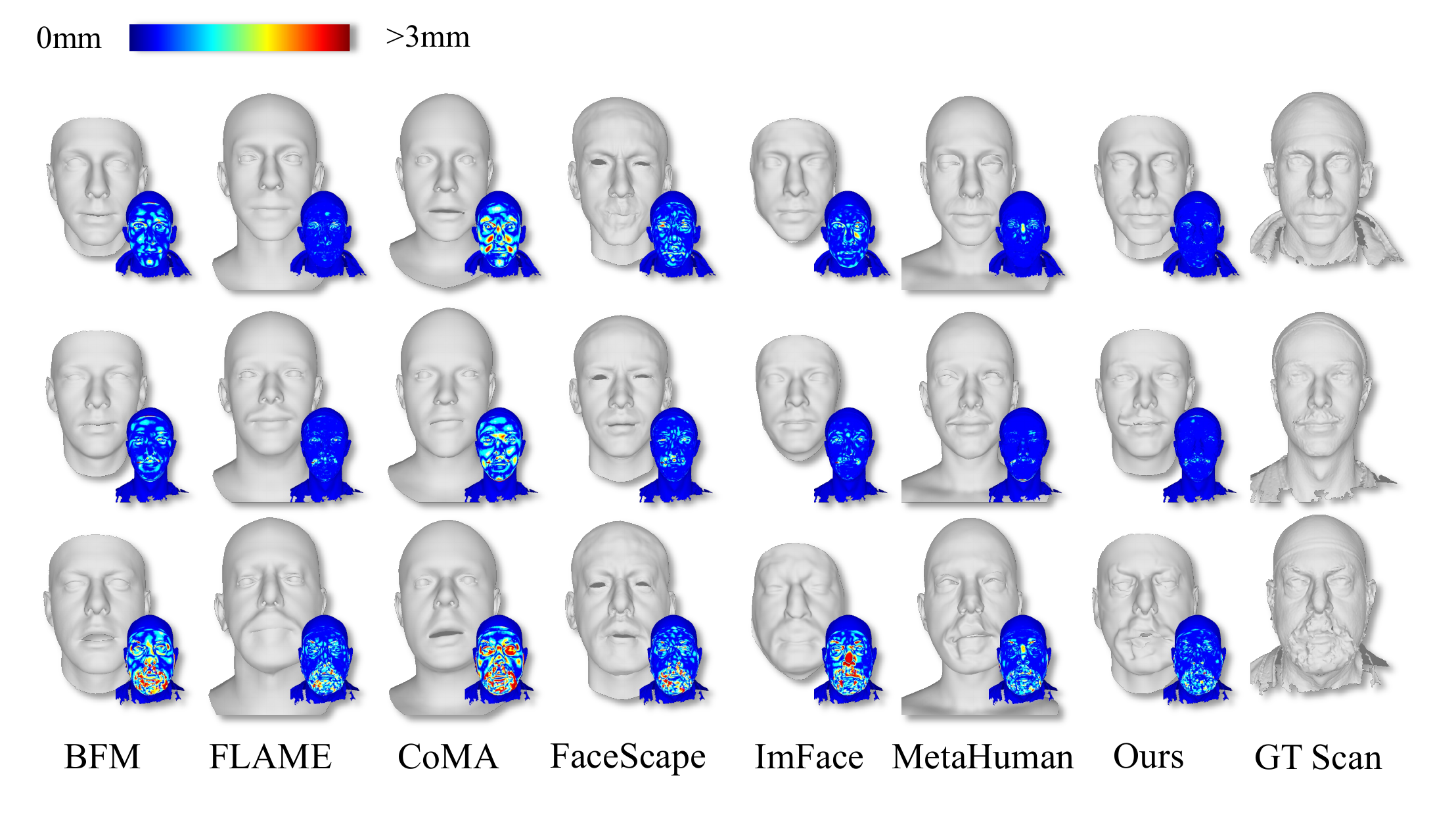}
	\end{center}
	\caption{Exemplar failed fitting results. GT Scans stands for the ground truth scan used for fitting. }
	\label{failed_cases}
\end{figure}

This study did not cover other aspects of multi-view reconstruction, such as the design of constraints or optimization process. We believe that our proposed model with higher capacity will facilitate future research on multi-view reconstruction, enabling better use of increased capacity to improve reconstruction performance. Another potential future work is to explore decoupling the identity and expression of the skinning parameters to enable expression transfer between different individuals and customization of personal-specific expressions. It would be interesting to combine data dependent decoupling techniques as used in 3DMMs with skinning models. Moreover, note that this study did not fully investigate the unique case of faces with beards. Given that beards are not part of the facial topology, they pose a significant challenge for all parametric face models. This is particularly true for models with higher capacity, such as ASM, which is more prone to artifacts, as illustrated in Fig.~\ref{failed_cases}.

\section{Conclusion}
We proposed ASM, a high-capacity parametric face model, to be used for reconstruction with multi-view uncalibrated images. ASM offers stronger capacity than data-dependent 3DMMs with compact and fully tunable parameters. Our experiments demonstrated that ASM achieved SOTA performance for multi-view reconstruction on the MICC Coop benchmark, and its high capacity was crucial to exploit abundant information from multi-view input. Furthermore, the semantic parameters of ASM made it suitable for real-world applications like in-game avatar creation. The study opens up new research direction for the parametric face model and facilitates future research on multi-view reconstruction.

\section{Acknowledgment}
We would like to thank Jiawen Zheng, Bishan Wang and Shaowen Xie for their assistance in the data collection process.

{\small
	\bibliographystyle{ieee_fullname}
	\bibliography{main}
}

\clearpage
\appendix

\section{Model Implementation Details}
\begin{figure}[htbp]
	\begin{center}
		\vspace{-0pt}
		\includegraphics[width=0.45\textwidth]{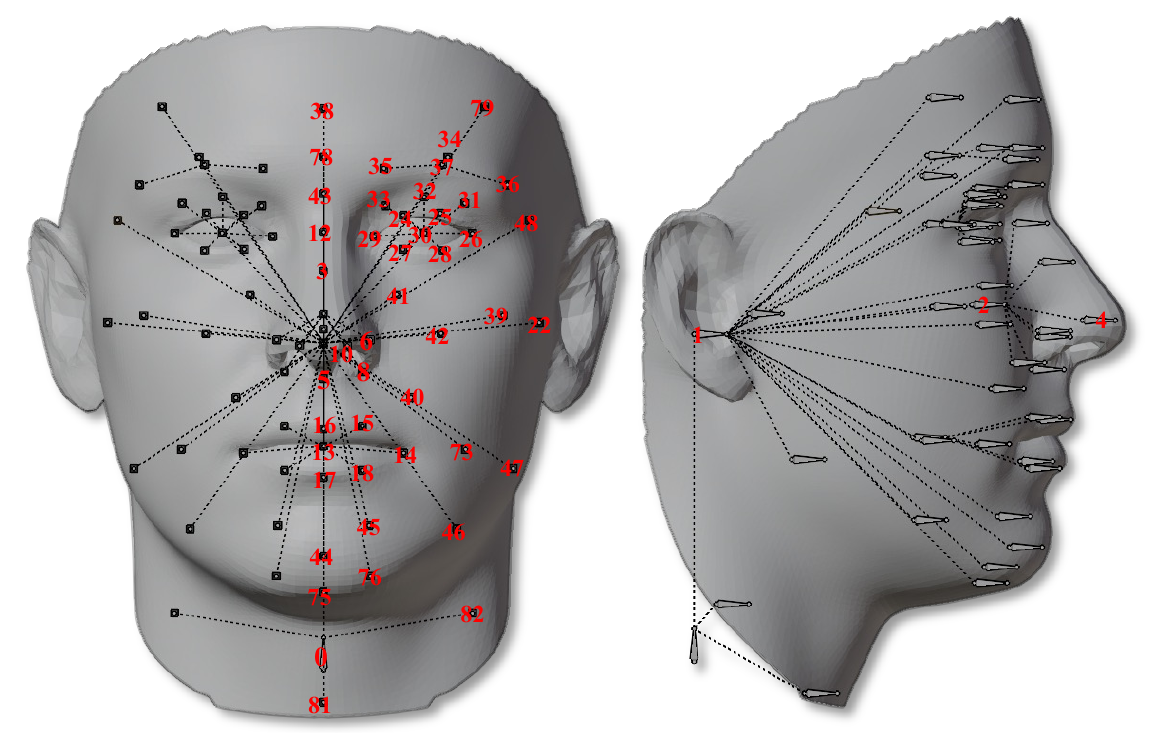}
	\end{center}
	\caption{Illustration of the initial binding of ASM. The index numbers of bones on the left half of the face are displayed, while the index numbers of the right counterparts are omitted. }
	\label{bone-binding-figure}
\end{figure}

We utilized Blender to place 84 bones on the face, as depicted in Fig.~\ref{bone-binding-figure}. The bone arrangement was derived from the distribution of the human skeletons and musculature to better represent the facial details of human faces. The orientations of all the bones are aligned parallel to the z-axis, except the root bone. The list of bone indices and names can be found in Tab.~\ref{bone-names}

\begin{table}[!htbp]
	\small
	\begin{center}
		\resizebox{0.46\textwidth}{!}{
			\addtolength{\tabcolsep}{-4pt}
			\small
			\begin{tabular}{cccccc}
				\toprule
				Indices & Parents & Names & Indices & Parents & Names\\
				\midrule
				0 & - & root & 42 & head & apple$\_$center.L \\
				1 & root & head  & 43 & head & eyebrow$\_$center \\
				2& head & nose & 44 & head & chin \\
				3 & nose & nose$\_$bridge & 45 & head & chin$\_$side.L \\
				4 & nose & nose$\_$tip & 46 & head & jaw.L \\
				5 & nose & nose$\_$mid & 47 & head & jaw$\_$corner.L \\
				6 & nose & nose$\_$wing.L & 48 & head & temple.L \\
				7 & nose & nose$\_$wing.R & 49 & head & ear.R \\
				8 & nose & nose$\_$bottom.L & 50 & head & eye.R \\
				9 & nose & nose$\_$bottom.R & 51 & eye.R & eye$\_$inner$\_$upper.R \\
				10 & nose & nose$\_$hole.L & 52 & eye.R & eye$\_$outer$\_$upper.R \\
				11 & nose & nose$\_$hole.R & 53 & eye.R & eye$\_$outer$\_$corner.R \\
				12 & nose & nose$\_$bridge$\_$upper & 54 & eye.R & eye$\_$inner$\_$lower.R \\
				13 & head & mouth & 55 & eye.R & eye$\_$outer$\_$lower.R \\
				14 & mouth & lip$\_$corner.L & 56 & eye.R & eye$\_$inner$\_$corner.R \\
				15 & mouth & lip$\_$upper$\_$side.L & 57 & eye.R & eye$\_$hole.R \\
				16 & mouth & lip$\_$upper$\_$mid & 58 & eye.R & eyelid$\_$outer.R \\
				17 & mouth & lip$\_$lower$\_$mid & 59 & eye.R & eyelid$\_$middle.R \\
				18 & mouth & lip$\_$lower$\_$side.L & 60 & eye.R & eyelid$\_$inner.R \\
				19 & mouth & lip$\_$corner.L & 61 & head & eyebrow.R \\
				20 & mouth & lip$\_$upper$\_$side.R & 62 & eyebrow.R & eyebrow$\_$inner.R \\
				21 & mouth & lip$\_$lower$\_$side.R & 63 & eyebrow.R & eyebrow$\_$outer.R \\
				22 & head & ear.L & 64 & eyebrow.R & eyebrow$\_$mid.R \\
				23 & head & eye.L & 65 & head & apple$\_$outer.R \\
				24 & eye.L & eye$\_$inner$\_$upper.L & 66 & head & apple$\_$lower.R \\
				25 & eye.L & eye$\_$outer$\_$upper.L & 67 & head & apple$\_$inner.R \\
				26 & eye.L & eye$\_$outer$\_$corner.L & 68 & head & apple$\_$center.R \\
				27 & eye.L & eye$\_$inner$\_$lower.L & 69 & head & chin$\_$side.R \\
				28 & eye.L & eye$\_$outer$\_$lower.L & 70 & head & jaw.R \\
				29 & eye.L & eye$\_$inner$\_$corner.L  & 71 & head & jaw$\_$corner.R \\
				30 & eye.L & eye$\_$hole.L & 72 & head & temple.R \\
				31 & eye.L & eyelid$\_$outer.L & 73 & head & cheek.L \\
				32 & eye.L & eyelid$\_$middle.L & 74 & head & cheek.R \\
				33 & eye.L & eyelid$\_$inner.L & 75 & head & chin$\_$low \\
				34 & head & eyebrow.L & 76 & head & chin$\_$side$\_$low.L \\
				35 & eyebrow.L & eyebrow$\_$inner.L & 77 & head & chin$\_$side$\_$low.R \\
				36 & eyebrow.L & eyebrow$\_$outer.L & 78 & head & eyebrow$\_$center$\_$up \\
				37 & eyebrow.L & eyebrow$\_$mid.L & 79 & head & forehead.L \\
				38 & head & forehead & 80 & head & forehead.R \\
				39 & head & apple$\_$outer.L & 81 & root & neck$\_$front \\
				40 & head & apple$\_$lower.L & 82 & root & neck$\_$side.L \\
				41 & head & apple$\_$inner.L & 83 & root & neck$\_$side.R \\
				\bottomrule
		\end{tabular}}
	\end{center}
	\caption{Skeleton structure of ASM.}
	\label{bone-names}
\end{table}

\section{Dynamic Bone Binding}
In the section of the main text on Dynamic Bone Binding, we discussed the process of updating the entire bone binding by modifying the UV coordinates of the bones. In the following section, we will provide additional elaboration on this topic.

\hspace*{\fill}

{\noindent \bf Barycentric Interpolation.} 
We check whether the UV coordinate of bone $j$, denoted as $\bm{\zeta}$, falls within triangle $f_{ABC}$ on UV space using the following formula:
\begin{equation}
	\begin{aligned}
		\begin{split}
			&t1 = \overrightarrow{\zeta A} \times \overrightarrow{\zeta B} \\
			&t2 = \overrightarrow{\zeta B} \times \overrightarrow{\zeta C}\\
			&t3 = \overrightarrow{\zeta C} \times \overrightarrow{\zeta A}
		\end{split}
	\end{aligned}
	\label{triangle}
\end{equation}
where $\times$ represents the cross-product operation. Point $\bm{\zeta}$ lies inside the triangle $f_{ABC}$ when $t1$, $t2$, and $t3$ have the same sign ($t1 > 0$, $t2 > 0$, $t3 > 0$ or $t1 < 0$, $t2 < 0$, $t3 < 0$). We calculate the barycentric weights $\alpha$, $\beta$, and $\gamma$ of point $\bm{\zeta}$ within $f_{ABC}$ with:
\begin{equation}
	\alpha, \beta, \gamma = Barycentric(\bm{\zeta}, A, B, C)
	\label{barycentric}
\end{equation}
where 
\begin{equation}
	\begin{split}
		\begin{aligned}
			&\alpha = \frac{-(x_\zeta - x_B)(y_C - y_B) + (y_\zeta - y_B)(x_C- x_B)}{-(x_A - x_B)(y_C - y_B) + (y_A - y_B)(x_C - x_B)} \\
			&\beta = \frac{-(x_\zeta - x_C)(y_A- y_C) + (y_\zeta - y_C)(x_A- x_C)}{-(x_B - x_C)(y_A - y_C) + (y_B - y_C)(x_A- x_C)} \\
			&\gamma = 1 - \alpha - \beta
			\label{f5}
		\end{aligned}
	\end{split}
\end{equation}
After obtaining the values of $\alpha$, $\beta$, and $\gamma$, we use them to interpolate the coordinates of $\bm{\psi}$ in the world space. This is done based on the 3D coordinates of the three vertices, $\bm{v}_A$, $\bm{v}_B$, and $\bm{v}_C$. To recalculate $\bm{\psi}_j$ for bone $j$, we use the following formula:
\begin{equation}
	\bm{\psi}_j = F'(\bm{\zeta}_j) = \alpha \mathbf{v}_A + \beta \mathbf{v}_B + \gamma \mathbf{v}_C - \mathbf{v}_{t} + \bm{\psi}_j^0
	\label{psi}
\end{equation}
where $\bm{\zeta}_j$ represents the UV coordinates, $\mathbf{v}_t$ represents the vertex closest to $\bm{\psi}_j^0$, and $\bm{\psi}_j^0$ represents the initial position of bone $j$. 

\hspace*{\fill}

{\noindent \bf Binding Updated.} 
We provide a brief overview of the Linear Blend Skinning (LBS) method:
\begin{equation}
	\mathbf{v}' = \sum^J_{j=1} w_j \mathbf{T}_j \mathbf{v}
	\label{lbs2}
\end{equation}
The deformation is achieved through the use of $w_j$ and $\mathbf{T}_j$, as demonstrated by Eq.~\ref{lbs2}. We expand $\mathbf{T}_j$ according to the following formula:
\begin{equation}
	\begin{split}
		\begin{aligned}
			\mathbf{T}_j &= \mathbf{M}^{l2w}_j \mathbf{M}^{w2l}_j \\
			& = \mathbf{M}^{l2w}_j \mathbf{B}_j^{-1}  \\
			& = \mathbf{M}^{l2w}_p\mathbf{M}^{trs}(\bm{\tau}_j)\mathbf{B}_j^{-1} \\ 
			& = \prod_{p=1}^{P}\mathbf{M}^{trs}(\bm{\tau}_p)\mathbf{M}^{trs}(\bm{\tau}_j)\mathbf{B}_j^{-1}
			\label{transform}
		\end{aligned}
	\end{split}
\end{equation}
where $\mathbf{B}_j$ represents the bind-pose of bone $j$, and $\mathbf{M}^{trs}(\cdot)$ will be described in detail below. $P$ denotes the parent chain for bone $j$. For example, the parent chain for nose$\_$tip represents (nose$\_$tip - nose - head - root). Eq.~\ref{transform} shows that $\mathbf{T}_j$ has two parts: the bind-pose matrix for bone $j$ that converts vertex $\mathbf{v}$ from the world space to the local space and the result of multiplying the transformation matrix of bone $j$ with the local-to-world matrix of its parent bone, which converts $\mathbf{v}$ from the local space to the world space.  When updating the 3D world position of bone $j$, it is necessary to update $\mathbf{B}_j$ and recalculate $\mathbf{M}^{trs}$ using the updated relative position between bones. The first step is to update the bind-pose matrix of each bone. 

We keep rotation and scaling constant as in the initial binding, and only the translation component of the bind-pose matrix needs to be updated:
\begin{equation}
	\mathbf{B}_j = B(\bm{\psi}_j) = 
	\begin{bmatrix}
		R_j^0S_j^0 & \bm{\psi}_j  \\
		0 & 1
	\end{bmatrix}
	\label{update_B}
\end{equation}
where $R_j^0$ and $S_j^0$ represent the rotation and scaling matrices identical to those present in the initial bind-pose matrix. $\bm{\psi}_j$ is the updated 3D world space position obtained from Eq.~\ref{psi}. Assuming that bones undergo just translation without rotation greatly simplifies the dynamic bone binding calculation stage. 

We simply recalculate the new bind-pose matrix from the updated world coordinates of each bone. In the following, we will introduce how to calculate the transformation matrix by taking into account the updated relative positions between bones.

We define $\bm{\tau} = \left[ \mathbf{r}^T, \mathbf{t}^T, \mathbf{s}^T \right]^T \in \mathbb{R}^9$ as the pose parameters in the local space of bone $j$, we decompose $\mathbf{M}^{trs}(\bm{\tau}_j)$ with:
\begin{equation}
	\mathbf{M}^{trs}(\bm{\tau}_j) = \mathbf{M}^{l2p}_j \mathbf{M}(\bm{\tau}_j)
	\label{M_trs}
\end{equation}
where $\mathbf{M}(\bm{\tau}_j)\in\mathbb{R}^{4\times4}$ is the standard transformation matrix composed from $\bm{\tau}_j$. $\mathbf{M}^{l2p}_j $ transform bone $j$ into the coordinate system of its parent bone $p$, which can be solely determined by the bind-pose of the bones:
\begin{equation}
	\begin{split}
		\begin{aligned}
			\mathbf{M}^{l2p}_j = \mathbf{B}_p^{-1}\mathbf{B}_j
		\end{aligned}
		\label{local_update}
	\end{split}
\end{equation}
where $ \mathbf{B}_p$ and $ \mathbf{B}_j$ are the bind-pose matrix of bone $p$ and bone $j$, respectively.

With the introduction of these concepts, we complete the dynamic adjustment process for bone binding with Eq.~\ref{transform}. For details on the Dynamic Bone Binding process, refer to Algorithm~\ref{alg1}.

\setlength{\textfloatsep}{10pt}
\begin{algorithm}
	\renewcommand{\algorithmicrequire}{\textbf{Input:}}
	\renewcommand{\algorithmicensure}{\textbf{Output:}}
	\caption{Dynamic Bone Binding}
	\label{alg1}
	\begin{algorithmic}[1]
		\REQUIRE $\bm{\zeta}, \bm{\tau}, \bm{\psi}^0, \mathbf{v_t}, \mathbf{B}^0$ 
		
		\tcp{barycentric interpolation phase}
		\FOR{bone $j$ in skeleton}
		\FOR{triangle $f$ in UV map}
		\IF{$\bm{\zeta}_j$ falls within $f$ based on Eq.~\ref{triangle}}
		\STATE Update $\alpha, \beta, \gamma$ with $f$ based on Eq.~\ref{barycentric}
		\STATE Update $\bm{\psi}_j$ with $\alpha, \beta, \gamma$ based on Eq.~\ref{psi}
		\ENDIF
		\ENDFOR
		\ENDFOR
		
		\hspace*{\fill} 
		
		\tcp{binding updated phase}
		\FOR{bone $j$ in skeleton}
		\STATE Update $\mathbf{B}_j$ with $\bm{\psi}_j$ and $\mathbf{B}_j^0$ based on Eq.~\ref{update_B}
		\STATE Update $\mathbf{M}^{l2p}_j$ with $\mathbf{B}_j$ and $\mathbf{B}_p$ based on Eq.~\ref{local_update}
		\STATE Update $\mathbf{M}^{trs}(\bm{\tau}_j)$ with $\bm{\tau}_j$ and $\mathbf{M}^{l2p}_j$ based on Eq.~\ref{M_trs}
		\ENDFOR
		
		\FOR{bone $j$ in skeleton}
		\STATE Update $\mathbf{T}_j$ with $\mathbf{M}^{trs}(\bm{\tau}_j)$, $\prod_{p=1}^{P}\mathbf{M}^{trs}(\bm{\tau}_p)$ and $\mathbf{B}_j$ based on Eq.~\ref{transform}
		\ENDFOR
		
		\ENSURE $\mathbf{T}$ 
	\end{algorithmic}  
\end{algorithm}

\section{Face Registration} \label{face-registration}
We conducted face registration experiments on LYHM~\cite{dai2020statistical} and FaceScape~\cite{yang2020facescape} to evaluate the representation capacity of different parametric face models. We marked 7 key points on the ground-truth scans, consistent with the NoW-Benchmark prototype~\cite{sanyal2019learning}. We identified an equivalent number of key points on the model's topology and utilized the 3D coordinates of 7 key point pairs to compute global rigid transformation parameters as initial values for fitting. We used the $point\_mesh\_face\_distance()$ function from PyTorch3D to optimize the distance between the ground-truth scan vertices and the nearest triangle on the predicted mesh. We compared our model with BFM~\cite{paysan20093d}, FLAME~\cite{li2017learning}, CoMA~\cite{ranjan2018generating} , FaceScape~\cite{yang2020facescape}, ImFace~\cite{zheng2022imface}, and MetaHuman~\cite{games2021metahuman}. 

BFM uses a 199-dimensional shape basis and a 79-dimensional expression basis in its optimization process. FLAME employs a 300-dimensional shape basis, a 100-dimensional expression basis, and rotation parameters for two joints, neck and chin, total of 406 parameters. We retrained the encoder and decoder networks of CoMA with 64-dimensional latent vectors on the CoMA datasets. During fitting, we only optimized the latent vector while fixing the decoder network parameters. FaceScape uses a 300-dimensional shape basis and a 51-dimensional expression basis, excluding the natural expression from the original 52-dimensional expression basis. To preserve the mesh's initial scale, we subtracted the sum of the remaining 51 dimensions to obtain the value of the natural expression basis. This approach ensured that the sum of all 52 expression dimensions equaled 1, thereby reducing one degree of freedom. For ImFace, we converted the ground truth mesh to SDF and fitted the corresponding $z\_id$ and $z\_exp$ using a pre-trained network. We then converted the SDF results to meshes and evaluated them in our metric-calculation prototype. For MetaHuman, we fitted all 9 degrees of freedom of the bones to deform the meshes.

We utilized Adam optimizer in PyTorch~\cite{paszke2019pytorch} to optimize the model parameters and the global rigid transformation parameters, minimizing the point-to-face distance, while adding regularization to avoid model artifacts. We kept the same learning rate of 1e-3 and iteration steps of 300 for all models, except for ImFace, for which we used the default iteration steps of 1,500 as used in the released code. 

\section{Multi-view Face Reconstruction}
Compared to Multi-view Stereo (MVS) without using the face priors, parametric face models greatly reduce the number of parameters to be solved, which alleviated the under-determined problem. We solved the multi-view face reconstruction with a parametric face model as an optimization problem with constraints from multiple views. For all the parametric face models in the experiments, we used the predicted results from a single-view face reconstruction network as the initial point for quick and stable convergence of the optimization process.

\subsection{Single-view Prediction Network}
We followed Deng~\etal~\cite{deng2019accurate}  and built a reconstruction network based on 3DMM. Given an image $I$ captured from one of the multiple views, we used a neural network to predict its parameter set $X=(\bm{\alpha}, \bm{\beta}, \mathbf{t},  \mathbf{R})$, where $\bm{\alpha}$ and $\bm{\beta}$ represented the shape and expression parameters of the 3DMM basis, respectively. The 3D face shape $\bm{S}$ could be represented by $\mathbf{S} = \bar{\mathbf{S}} + \mathbf{B}_{id}\bm{\alpha} + \mathbf{B}_{exp}\bm{\beta}$. The translation $\mathbf{t} \in \mathbb{R}^3$ and rotation $\mathbf{R} \in SO(3)$ represented the pose of the 3D model with respect to the current camera position. We used a perspective projection model to obtain the projected coordinates $\mathbf{p}$ of the vertices $\mathbf{v}$ in image $I$ as $\mathbf{p} = \Pi(\mathbf{Rv} + \mathbf{t})$, where $\Pi(\cdot)$ represented the perspective projection.

To obtain an initial shape for multi-view optimization, we simply averaged the predicted $\bm{\alpha}$ and $\bm{\beta}$ across multiple views to obtain the initial face mesh.
\begin{equation}
	\mathbf{S} = \bar{\mathbf{S}} + \mathbf{B}_{id}\bm{\alpha}_{mean} + \mathbf{B}_{exp}\bm{\beta}_{mean}
	\label{f0}
\end{equation}

To use the predicted face shape, an additional registration step is required to get initial model parameters to fit the predicted mesh. For BFM and ASM with the same topology as Deng~\etal~\cite{deng2019accurate}, the point-to-point loss was employed. For FLAME and MetaHuman with different topologies, point-to-face loss as described in Sec.~\ref{face-registration} was used. For all the models, the registration step was implemented with Adam optimizer with a learning rate of 1e-3 and iteration steps of 300. These model parameters, together with pose parameters, would be solved in the following optimization step for multi-view reconstruction. 

\subsection{Energy Function}

{\noindent \bf Landmarks.} 
We used the facial key points prediction method~\cite{bulat2017far} to predict the positions of 68 facial key points from the input image. We annotated the corresponding 68 vertices on the mesh and obtained the projected coordinates of these vertices on the corresponding image with the perspective projection function. We then minimized the landmarks term:
\begin{equation}
	E_{lmk}(\mathbf{\hat{v}}) = \frac{1}{N}\sum_{n=1}^{N} || \mathbf{q}^n - \Pi(\mathbf{R^n}\mathbf{\hat{v}} + \mathbf{t}^n)||^2
	\label{f8}
\end{equation}
where $\mathbf{q}^n $ were the key points from landmarks prediction, and $\mathbf{\hat{v}}$ were the masked 68 vertices.

\hspace*{\fill}

{\noindent \bf Regularization.} 
We imposed constraints on the parameters to be optimized in the parametric face models to prevent over-fitting.  For the ASM model parameters $\bm{\tau}, \bm{\zeta}, \bm{\pi}, \bm{\mu}, \bm{\Sigma}$, we used the following shape priors:
\begin{equation}
	\begin{split}
		\begin{aligned}
			E_{reg}(\bm{\tau}, \bm{\zeta}, \bm{\pi}, \bm{\mu}, \bm{\Sigma}) = &\lambda_1||\bm{\tau}||^2 + \lambda_2|| \bm{\zeta} ||^2 + \lambda_3|| \bm{\pi} -  \bm{\pi}'||^2 + \\ &\lambda_4|| \bm{\mu} -  \bm{\mu}'||^2 + \lambda_5|| \bm{\Sigma} -  \bm{\Sigma}'||^2
			\label{f9}
		\end{aligned}
	\end{split}
\end{equation}
where $\bm{\pi}', \bm{\mu}', \bm{\Sigma}'$ were the initial parameters of ASM from fitting the skinning weights generated by Blender. Here we use $\lambda_1 = 1, \lambda_2 = \lambda_3 = \lambda_4 = \lambda_5 = 0.1$

\hspace*{\fill}

{\noindent \bf Edge.} 
Due to the recent development of deep learning-based face parsing algorithms, we could easily obtain a refined segmentation of facial semantic area. We trained a face segmentation network on the CelebaMask-HQ datasets, and with an image $I$ we predicted its segmentation mask and extracted the face contour points set donated as $E$ for constraining the mesh's contour. 

To create the contour points set of the reconstructed face model, we applied a rigid transformation with the predicted pose parameters. We calculated the normal direction of each vertex on the model surface and computed the absolute value of the normal on the z-axis. We considered the z-axis absolute value of the vertices normal less than a threshold to be the points at the edge of the face model and included them in the candidate set. 

To ensure that only visible vertices were used, we utilized z-buffering to determine whether the vertices in the candidate point set are visible. This enabled us to screen out the final contour vertices. Finally, we minimized the edge term by extracting contour points from both the image and the model contour points set.
\begin{equation}
	\begin{split}
		\begin{aligned}
			&E_{edge}(\mathbf{v'}) = \frac{1}{N} \sum_{n=1}^{N} chamfer(E^n, {\Pi(\mathbf{\hat{\mathbf{v}}}^n)}) \\  &\hat{\mathbf{v}} = \{\mathbf{v}' | |N(\mathbf{v'})_z| < \theta \;and \; V(\mathbf{v'}) > 0 \} 
			\label{f10}
		\end{aligned}
	\end{split}
\end{equation}
where $\mathbf{v'} = \mathbf{R^n}\mathbf{v} + \mathbf{t}^n$ and $\theta$ was the threshold used for distinguishing edge vertices. $N(\cdot)$ was the normal calculation function for vertices and $V(\cdot)$ was the z-buffering test to decide whether vertex $\mathbf{v}$ was visible in the current frame, respectively. In experiments, we used $\theta$ of $10^{\circ}$.

\hspace*{\fill}

{\noindent \bf Photometric Consistency.} 
The photometric consistency constraint method has found wide application in solving the parameters of face models from multi-view images. The technique involves projecting the vertices of the mesh onto images from various angles and sampling the corresponding pixel intensities on the original image using the projection coordinates $\mathbf{p}$. Hernandez~\etal~\cite{hernandez2017accurate} utilizes a 3x3 intensity patch centered around $\mathbf{p}$ to use the photometric consistency. However, we have observed that this approach is not ideal when the original vertex $\mathbf{v}$ lies on a plane with significant depth variation. In such cases, the 3x3 patch includes pixel information with a broad depth range, resulting in less accurate photometric consistency constraints.

To address this limitation, we projected and sampled the intensity values for each vertex $\mathbf{v}_j$ on the model and obtain $S_j = \Gamma(\Pi(\mathbf{R}\mathbf{v}_j + \mathbf{t}))$,  where $\Gamma(\cdot)$ was the interpolation function for sample the intensity on the projection coordinate of $\mathbf{v}_j$. We then unwrapped the intensities of each vertex to the UV space to obtain the UV-intensities map $\mathbf{U} = {Unwrap} (\mathbf{S})$. 

In the UV space, we performed Local Normalized Cross-correlation (LNCC) on the UV map, 
\begin{equation}
	\begin{split}
		\begin{aligned}
			&LNCC(\mathbf{U^i}, \mathbf{U^j}) \\
			& = \frac{1}{X} \sum_{x=1}^{X} V(S^i_x) V(S^j_x) NCC(P(U^i_x), P(U^j_x))
			\label{f12}
		\end{aligned}
	\end{split}
\end{equation}
where $V(S^i_x) V(S^j_x) $ were the same z-buffering test function to decide the visible mask of points $S^i_x$ and $S^j_x$, respectively. $X$ were the total number of vertices. $P(U^i_x)$ is the 3x3 sampling function around pixel $U^i_x$, and $NCC$ was the Normalized Cross-correlation function. The LNCC function captured the photometric information on the UV space rather than the image space. This process generated a sliding window that conformed to the mesh surface of the model, accurately sampled the vertex intensities even in areas with significant depth variation, and reduced the depth of field effect caused by sampling at a large scale. 

We calculated the LNCC loss on the unwrapped UV maps between any two frames in the input multi-view images. We then minimized the photometric-consistency term:
\begin{equation}
	\begin{split}
		\begin{aligned}
			E_{pc}(\mathbf{v}) = \frac{2}{N(N-1)} \sum_{i=1}^{N} \sum_{j=i+1}^{N} LNCC(\mathbf{U}^i, \mathbf{U}^j)
			\label{f14}
		\end{aligned}
	\end{split}
\end{equation}

\hspace*{\fill}

\subsection{Multi-view Optimization} 
We finally sum up all the energy terms and optimized the parameters of ASM as:
\begin{equation}
	\setlength{\belowdisplayskip}{10pt}
	E_{total} = \lambda_1 * E_{lmk} + \lambda_2 * E_{edge} + \lambda_3 * E_{pc} + \lambda_4 * E_{reg} 
	\label{f15}
\end{equation}
where $\lambda_1 = 0.001$, $\lambda_2 = 0.4$, $\lambda_3 = 100$ and $\lambda_4 = 1$ in our experiments. We utilized Adam optimizer to optimize the parameters of the model and the pose, minimizing the energy term of Eq.~\ref{f15}. The same learning rate of 1e-4 and iteration steps of 500 were used for all the models.

\begin{figure*}[htbp]
	\begin{center}
		\includegraphics[width=1\textwidth]{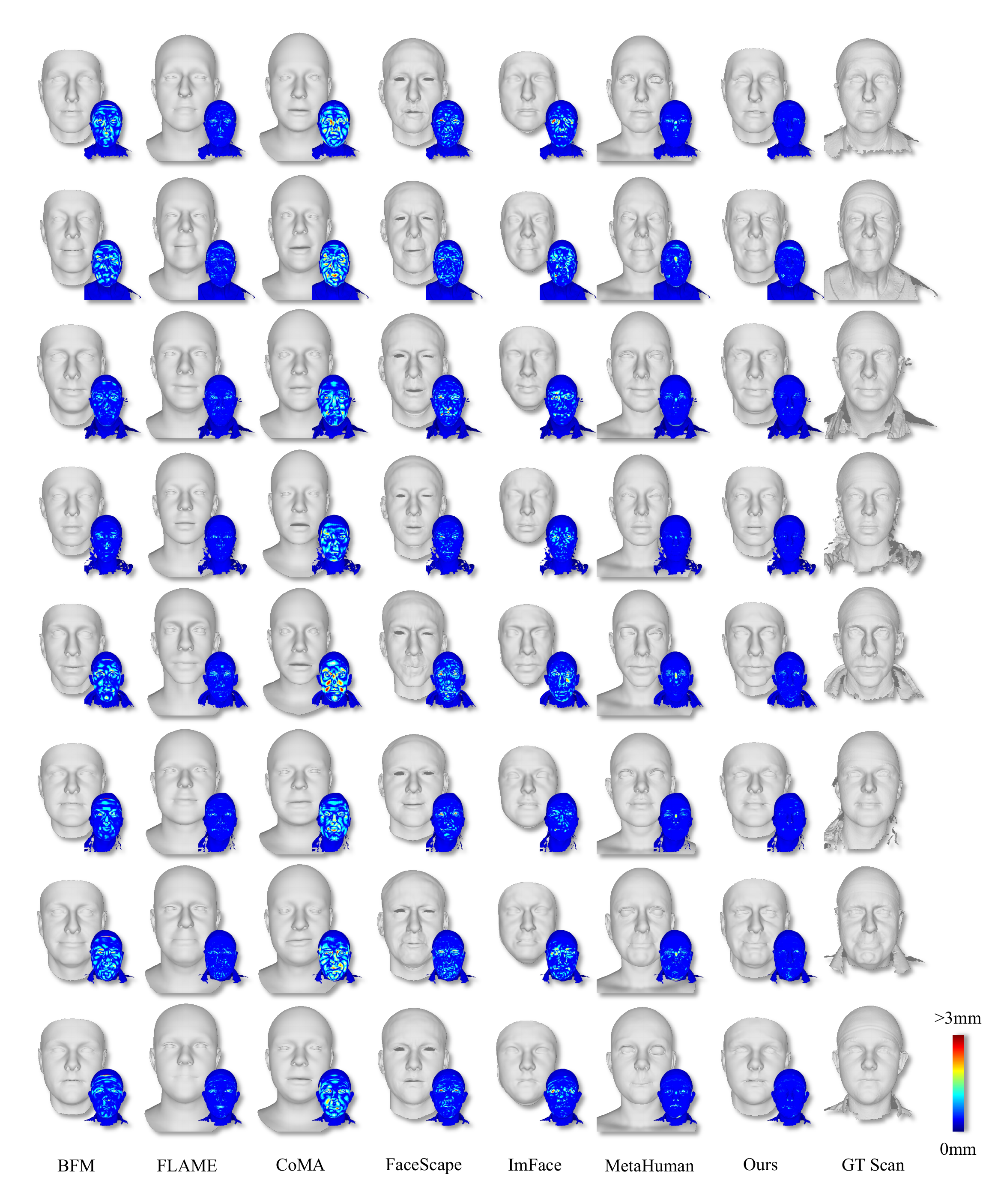}
	\end{center}
	\caption{More examples of fitting result on LYHM. GT Scans stand for the ground truth scan used for fitting.}
	\label{mc1}
\end{figure*}

\begin{figure*}[htbp]
	\begin{center}
		\includegraphics[width=1\textwidth]{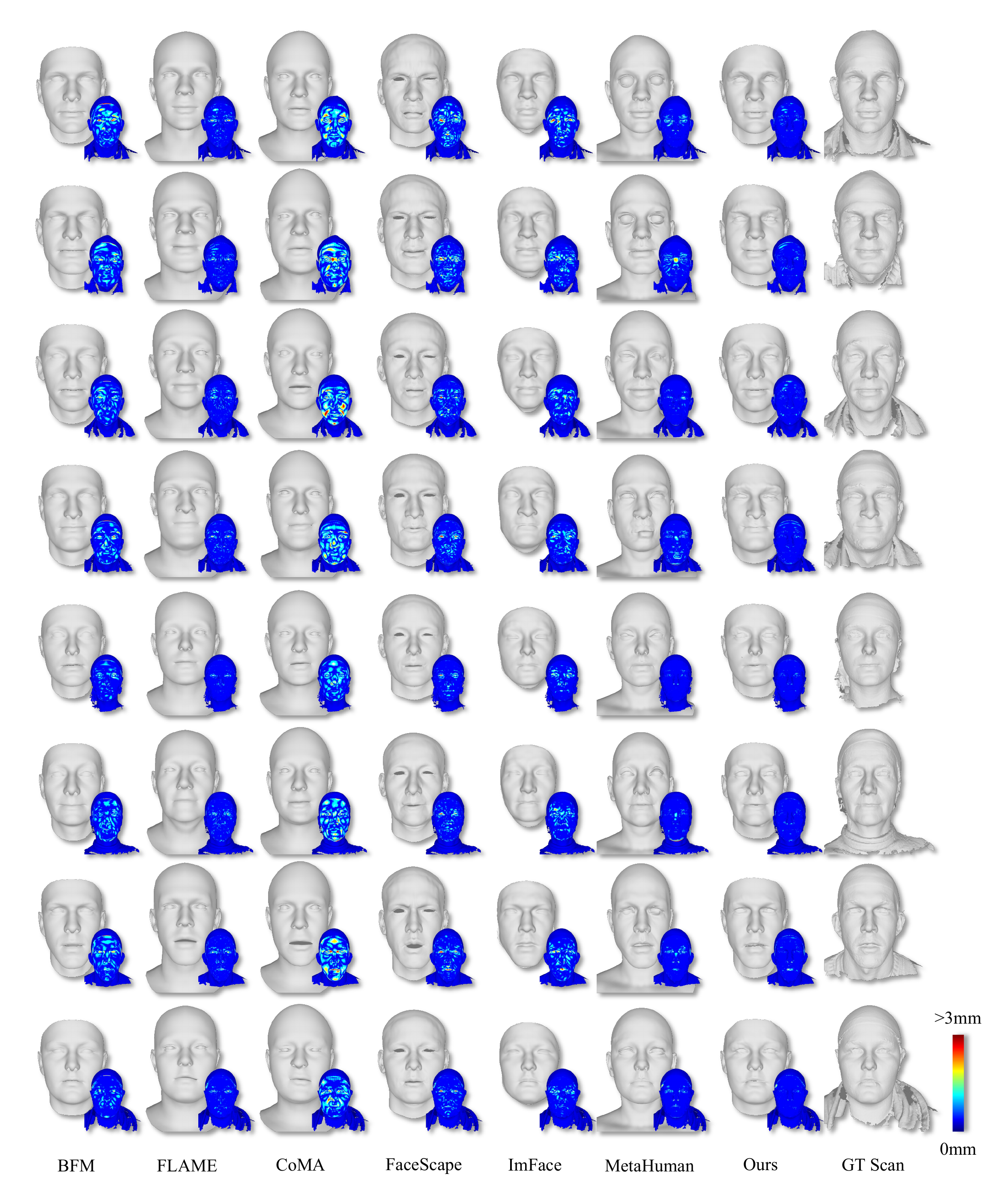}
	\end{center}
	\caption{More examples of fitting result on LYHM GT Scans stand for the ground truth scan used for fitting.}
	\label{mc2}
\end{figure*}

\begin{figure*}[htbp]
	\begin{center}
		\includegraphics[width=1\textwidth]{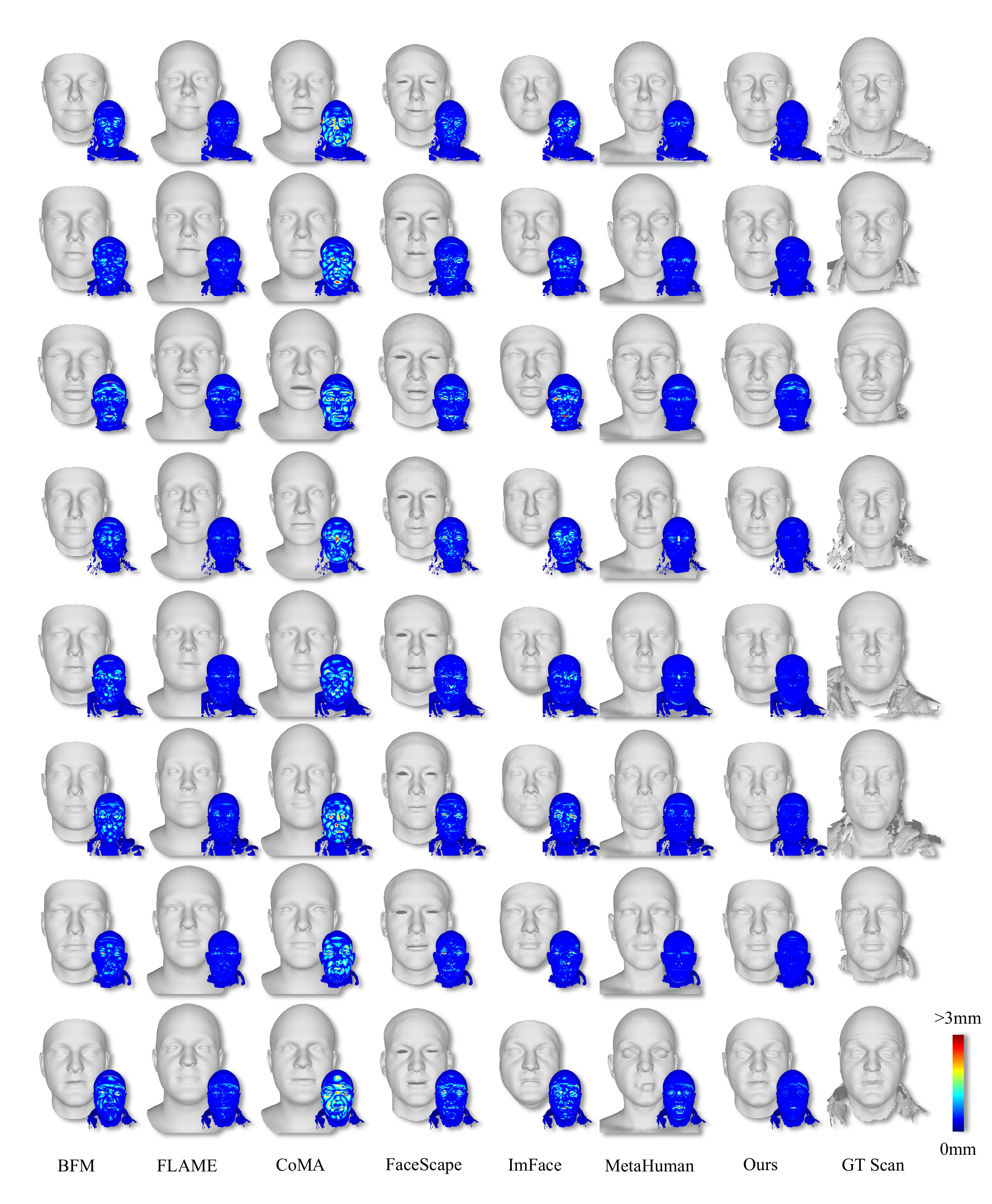}
	\end{center}
	\caption{More examples of fitting result on LYHM. GT Scans stand for the ground truth scan used for fitting.}
	\label{mc3}
\end{figure*}

\begin{figure*}[htbp]
	\begin{center}
		\includegraphics[width=1\textwidth]{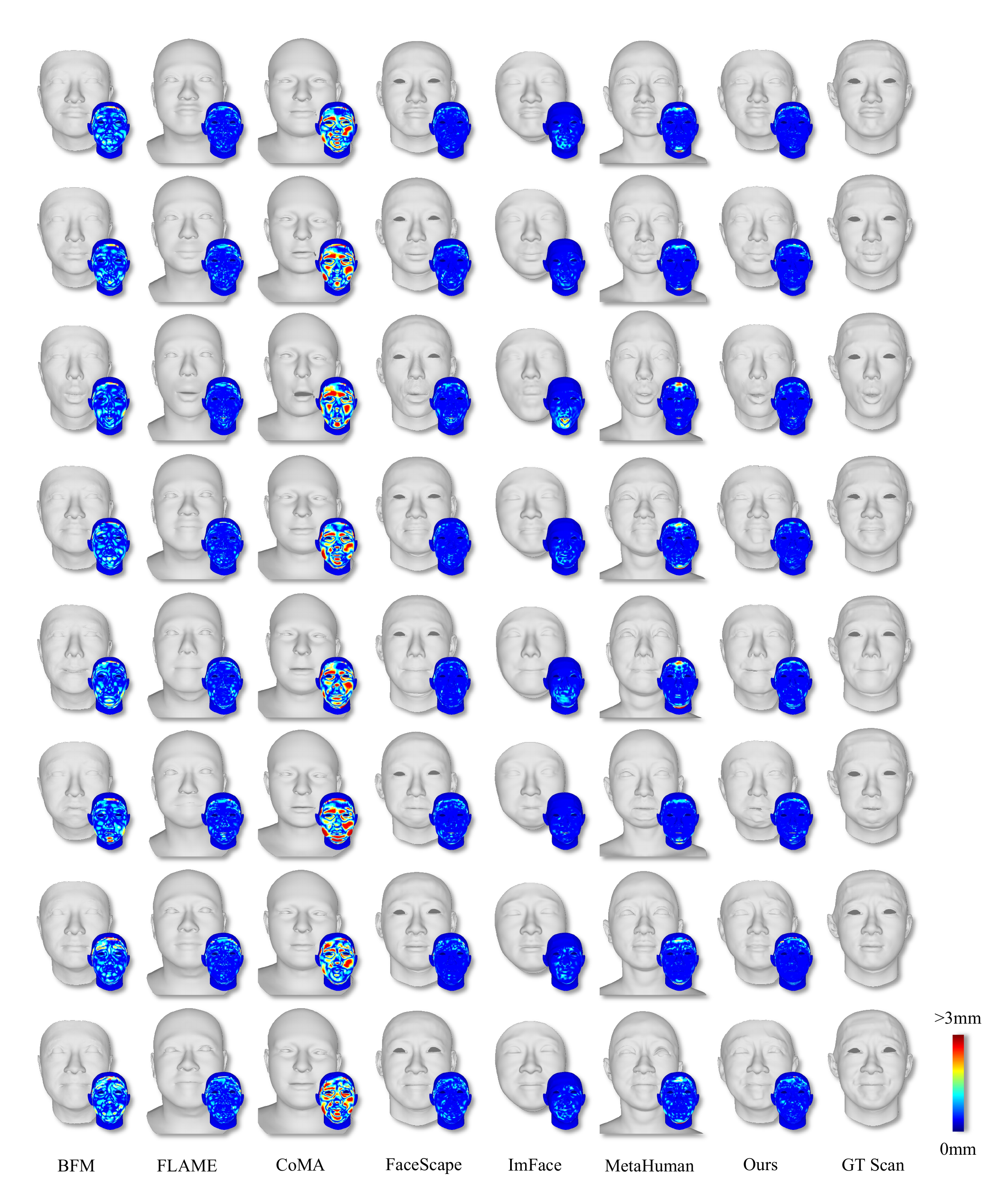}
	\end{center}
	\caption{More examples of fitting result on FaceScape. GT Scans stand for the ground truth scan used for fitting.}
	\label{mc4}
\end{figure*}

\end{document}